\PassOptionsToPackage{dvipsnames}{xcolor}
\documentclass[10pt,twocolumn,letterpaper]{article}

\usepackage{iccv}
\usepackage{times}
\usepackage{epsfig}
\usepackage{graphicx}
\usepackage{amsmath}
\usepackage{amssymb}
\usepackage{booktabs}
\usepackage{xcolor}
\usepackage{color}
\usepackage{colortbl}
\usepackage{enumitem}
\usepackage{multirow}
\usepackage{xr-hyper}
\usepackage[accsupp]{axessibility} 
\usepackage{subcaption}

\usepackage[pagebackref=true,breaklinks=true,letterpaper=true,colorlinks,bookmarks=false,citecolor=Periwinkle]{hyperref}
\hypersetup{
    colorlinks=true,
    linkcolor=BrickRed,
    filecolor=Fuchsia,      
    urlcolor=Orchid,
    pdftitle={PoseFix: Correcting 3D Human Poses with Natural Language},
    }

\iccvfinalcopy

\newcommand{\myparagraph}[1]{\vspace{0.05cm}\noindent\textbf{#1}}
\definecolor{PaleOrange}{rgb}{0.988,0.898,0.804}
\newcolumntype{o}{>{\columncolor{PaleOrange}}c}
\definecolor{myrowcolor}{rgb}{0.9, 0.9, 0.9}

\begin{document}

\newcommand{\GA}[1]{{\color{violet}#1}}
\newcommand{\GPM}[1]{{\color{blue} GPM:#1}} 

\newcommand{\mA}{\mathcal{A}}
\newcommand{\mF}{\mathcal{F}}
\newcommand{\mI}{\mathcal{I}}
\newcommand{\mK}{\mathcal{K}}
\newcommand{\mP}{\mathcal{P}}
\newcommand{\mR}{\mathcal{R}}
\newcommand{\mU}{\mathcal{U}}
\newcommand{\mZ}{\mathcal{Z}}

\newcommand{\bA}{\mathbf{A}}
\newcommand{\bB}{\mathbf{B}}
\newcommand{\bC}{\mathbf{C}}
\newcommand{\bD}{\mathbf{D}}
\newcommand{\bF}{\mathbf{F}}
\newcommand{\btF}{\tilde{\mathbf{F}}}
\newcommand{\bG}{\mathbf{G}}
\newcommand{\bH}{\mathbf{H}}
\newcommand{\bI}{\mathbf{I}}
\newcommand{\bJ}{\mathbf{J}}
\newcommand{\bK}{\mathbf{K}}
\newcommand{\bL}{\mathbf{L}}
\newcommand{\bM}{\mathbf{M}}
\newcommand{\bN}{\mathbf{N}}
\newcommand{\bO}{\mathbf{O}}
\newcommand{\bP}{\mathbf{P}}
\newcommand{\btP}{\tilde{\mathbf{P}}}
\newcommand{\bQ}{\mathbf{Q}}
\newcommand{\bR}{\mathbf{R}}
\newcommand{\btR}{\tilde{\mathbf{R}}}
\newcommand{\btS}{\tilde{\mathbf{S}}}
\newcommand{\bS}{\mathbf{S}}
\newcommand{\bT}{\mathbf{T}}
\newcommand{\btT}{\tilde{\mathbf{T}}}
\newcommand{\bU}{\mathbf{U}}
\newcommand{\bV}{\mathbf{V}}
\newcommand{\btV}{\tilde{\mathbf{V}}}
\newcommand{\bW}{\mathbf{W}}
\newcommand{\bX}{\mathbf{X}}
\newcommand{\btX}{\tilde{\mathbf{X}}}
\newcommand{\bY}{\mathbf{Y}}
\newcommand{\btY}{\tilde{\mathbf{Y}}}
\newcommand{\bZ}{\mathbf{Z}}

\newcommand{\bzero}{\textbf{0}}
\newcommand{\bone}{\textbf{1}}
\newcommand{\ba}{\mathbf{a}}
\newcommand{\bb}{\mathbf{b}}
\newcommand{\bc}{\mathbf{c}}
\newcommand{\bd}{\mathbf{d}}
\newcommand{\be}{\mathbf{e}}
\newcommand{\bff}{\mathbf{f}}
\newcommand{\bg}{\mathbf{g}}
\newcommand{\bh}{\mathbf{h}}
\newcommand{\bi}{\mathbf{i}}
\newcommand{\bj}{\mathbf{j}}
\newcommand{\bl}{\mathbf{l}}
\newcommand{\bm}{\mathbf{m}}
\newcommand{\bn}{\mathbf{n}}
\newcommand{\bo}{\mathbf{o}}
\newcommand{\bp}{\mathbf{p}}
\newcommand{\bq}{\mathbf{q}}
\newcommand{\br}{\mathbf{r}}
\newcommand{\bs}{\mathbf{s}}
\newcommand{\bts}{\tilde{\mathbf{s}}}
\newcommand{\bt}{\mathbf{t}}
\newcommand{\bu}{\mathbf{u}}
\newcommand{\btu}{\tilde{\mathbf{u}}}
\newcommand{\bv}{\mathbf{v}}
\newcommand{\btv}{\tilde{\mathbf{v}}}
\newcommand{\bbv}{\bar{\mathbf{v}}}
\newcommand{\bw}{\mathbf{w}}
\newcommand{\bx}{\mathbf{x}}
\newcommand{\btx}{\tilde{\mathbf{x}}}
\newcommand{\by}{\mathbf{y}}
\newcommand{\bty}{\tilde{\mathbf{y}}}
\newcommand{\bz}{\mathbf{z}}
\newcommand{\btz}{\tilde{\mathbf{z}}}
\newcommand{\bhz}{\hat{\mathbf{z}}}

\renewcommand{\vec}[1]{\boldsymbol{#1}}
\newcommand{\mat}[1]{\mathbf{#1}}
\newcommand{\set}[1]{\mathcal{#1}}

\newcommand{\real}[0]{\mathbb{R}}
\newcommand{\tb}[0]{\textbf}
\newcommand{\ti}[0]{\textit}
\newcommand{\et}[0]{\ti{et al.}}

\newcommand{\imageset}[0]{\set{I}}
\newcommand{\image}[0]{\mat{I}}

\newcommand{\poseset}[0]{\set{P}}
\newcommand{\transset}[0]{\set{T}}
\newcommand{\jointset}[0]{\set{J}}
\newcommand{\garmparamset}[0]{\set{G}}

\newcommand{\template}[0]{\mat{T}}
\newcommand{\garment}[0]{\mat{G}}
\newcommand{\blendweight}[0]{w}
\newcommand{\blendweights}[0]{\mat{W}}

\newcommand{\normal}[0]{{\mathbf{n}}}

\newcommand{\depth}[0]{\widehat{\vec{d}}}
\newcommand{\ankl}[0]{\mathbf{x}_{l}}
\newcommand{\ankr}[0]{\mathbf{x}_{r}}
\newcommand{\lplane}[0]{L_{p}}
\newcommand{\lamdp}[0]{\lambda_{p}}

\newcommand{\pose}[0]{\vec{\theta}}
\newcommand{\shape}[0]{\vec{\beta}}

\newcommand{\joints}[0]{\mat{J}}
\newcommand{\jointsTwoD}[0]{\tilde{\mat{J}}}

\newcommand{\rot}[0]{\vec{R}}

\newcommand{\scale}[0]{\mat{s}}
\newcommand{\trans}[0]{\vec{t}}

\newcommand{\nplane}[0]{\bf{\widehat{{n}}} }

\newcommand{\garmparam}[0]{\vec{z}}
\newcommand{\offsets}[0]{\mathbf{D}}

\newcommand{\cut}{\mat{z}_\mathrm{cut}}
\newcommand{\style}{\mat{z}_\mathrm{style}}
\newcommand{\posenc}{\mat{P}_{\shape}}
\newcommand{\zpose}{\mat{z}_{\pose}}

\newcommand{\smpl}[0]{M}
\newcommand{\posefun}[0]{T}
\newcommand{\blendfun}[0]{W}
\newcommand{\offsetfun}[0]{B}
\newcommand{\offsetsfun}[0]{D}
\newcommand{\jointfun}[0]{J}
\newcommand{\garmfun}[0]{G}

\newcommand{\metricdist}{pairwise normalized distances between persons}

\newcommand{\numberOfMetrics}{three} 

\newcommand{\Modelname}{Keep Your Feet on the Ground} 

\newcommand{\lossRep}[0]{L}
\newcommand{\multilossRep}[0]{\hat{L}}
\newcommand{\lambdaRep}[0]{\lambda_{2D}}

\newcommand{\heightDistMetric}[0]{h_{err}}

\newcommand{\posevect}[0]{\vec{p}}
\newcommand{\posevectRef}[0]{\vec{q}}
\newcommand{\jointsDims}[0]{\mathbb{R}^{J\times3}}
\newcommand{\diffVect}[0]{\vec{\delta}}

\newcommand{\featureVect}[0]{\vec{f}}
\newcommand{\featureNet}[0]{\mathcal{G}}
\newcommand{\ourNet}[0]{\Phi}
\newcommand{\rfineNet}[0]{\mathcal{R}}

\newcommand{\reals}[0]{\mathbb{R}}

\newcommand{\xmark}{\ding{55}}%

\newcommand{\bSigma}{\boldsymbol{\Sigma}}
\newcommand{\bpi}{\boldsymbol{\pi}}
\newcommand{\bmu}{\boldsymbol{\mu}}
\newcommand{\dname}{PoseFix\xspace}%

\title{\dname: Correcting 3D Human Poses with Natural Language}

\author{Ginger Delmas\textsuperscript{1,2}, Philippe Weinzaepfel\textsuperscript{2}, Francesc Moreno-Noguer\textsuperscript{1}, Gr\'egory Rogez\textsuperscript{2} \\[0.5cm]
\textsuperscript{1} Institut de Robòtica i Informàtica Industrial, CSIC-UPC, Barcelona, Spain\\
\textsuperscript{2} NAVER LABS Europe\\
\textsuperscript{1}{\tt\small \{gdelmas, fmoreno\}@iri.upc.edu, \textsuperscript{2}{\tt\small \{name.surname\}@naverlabs.com}} 
}

\maketitle


\begin{abstract}
Automatically producing instructions to modify one's posture could open the door to endless applications, such as personalized coaching and in-home physical therapy. Tackling the reverse problem (\ie, refining a 3D pose based on some natural language feedback) could help for assisted 3D character animation or robot teaching, for instance.
Although a few recent works explore the connections between natural language and 3D human pose, none focus on describing 3D body pose differences. In this paper, we tackle the problem of correcting 3D human poses with natural language.
To this end, we introduce the \dname dataset, which consists of several thousand paired 3D poses and their corresponding text feedback, that describe how the source pose needs to be modified to obtain the target pose. We demonstrate the potential of this dataset on two tasks: (1) {\em text-based pose editing}, that aims at generating  corrected 3D body poses given a query pose and a text modifier; and (2) {\em correctional text generation}, where instructions are generated based on the differences between two body poses. The dataset and the code are available at {\small \url{https://europe.naverlabs.com/research/computer-vision/posefix/}}. \normalsize
\end{abstract}


\section{Introduction}

How many puzzles could you solve with two human body poses and a description of their differences?
Call this description a feedback. It could be automatically generated by a fitness application based on the comparison between the gold standard fitness pose and the pose of John Doe, exercising in front of their smartphone camera in their living room (\textit{``straighten your back''}). In another context, the feedback can be considered a modifying instruction, provided by a digital animation artist to automatically modify the pose of a character, without having to redesign everything by hand. This feedback could be some kind of constraint, to be applied to a whole sequence of poses (make them run, but \textit{``with hands on the hips!''}). It could also be a hint, to guide pose estimation from images in failure cases: start from an initial 3D body pose fit, and give step-by-step instructions for the model to improve its pose estimation ({\textit{``the left elbow should be bent to the back''}).

\begin{figure}[t]
\centering
\includegraphics[width=\linewidth]{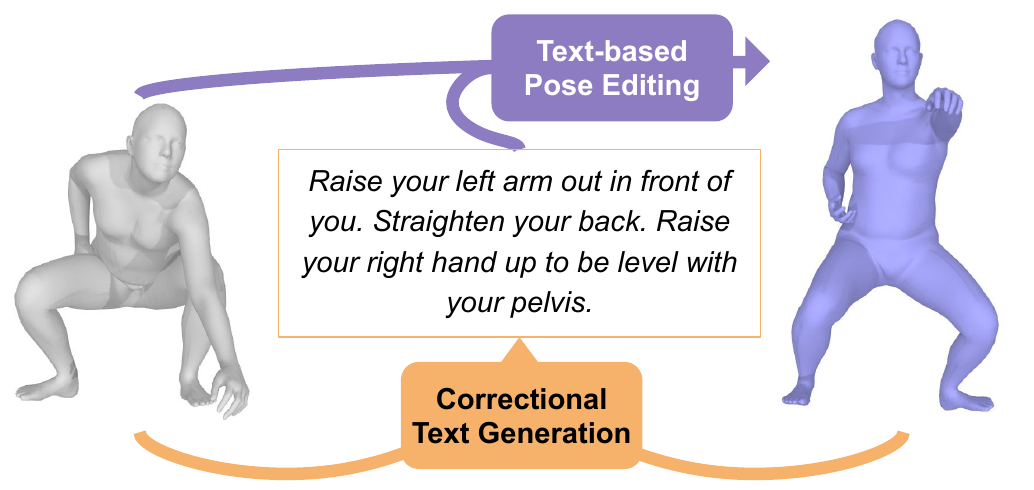} \\[-0.2cm]
\caption{\textbf{Illustration of the tasks addressed with the new \dname dataset}, which consists of textual descriptions of the difference between two 3D body poses.
}
\label{fig:front}
\vspace{-0.4cm}
\end{figure}

\begin{figure*}[t]
\centering
\includegraphics[width=\linewidth]{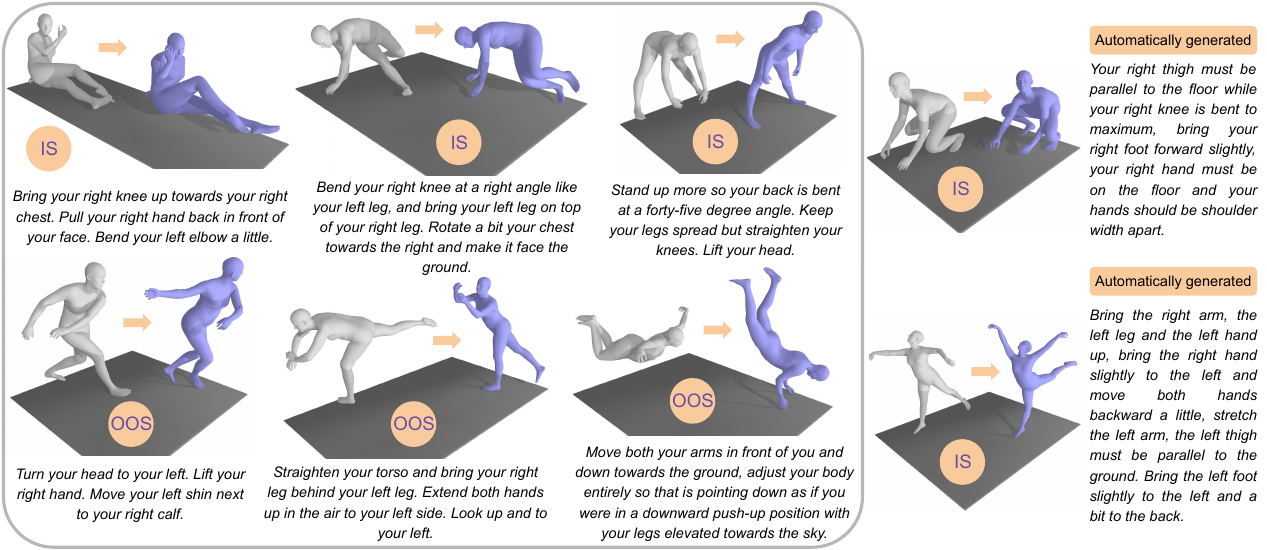} \\[-0.2cm]
\caption{\textbf{Examples of pose pairs and their annotated modifier in \dname.} The source pose is shown in gray and the target pose in purple. Poses from in-sequence (\textit{IS}) pairs are from the same motion clip; unlike out-of-sequence (\textit{OOS}) pairs.}
\label{fig:annotation_examples}
\vspace{-0.3cm}
\end{figure*}

In this paper, we focus on free-form feedback describing the change between two static 3D human poses (which can be extracted from actual pose sequences). Why so static? There exist many settings that require the semantic understanding of fine-grained changes of static body poses. For instance, yoga poses are extremely challenging and specific (with a lot of subtle variations), and they are static. Some sport motions require almost-perfect postures at every moment: for better efficiency, to avoid any pain or injury, or just for better rendering \eg in classical dance, yoga, karate, \etc
What is more, the realization of complex motions sometimes calls for precise step-to-step instructions, in order to assimilate the gesture or to perform it correctly.

Natural language can help in all these scenarios, in that it is highly semantic and unconstrained, in addition of being a very intuitive way to convey ideas. 
While 3D poses can be manually edited within a design framework~\cite{oreshkin2021protores}, language is particularly efficient for non-experts or when direct manipulation is not possible. The pose semantic we propose to learn here can be leveraged for other modalities (\eg images) or in other settings (\eg robot teaching).

While the link between language and images has been extensively studied in tasks like  image captioning~\cite{coco, herdade2019image} or image editing~\cite{Zhou_2019_CVPR}, the research on leveraging natural language for 3D human modeling is still in its infancy. A few works use textual descriptions to generate motion~\cite{Guo_2022_CVPR, tevet2022motionclip}, to describe the difference in poses from   synthetic \textit{2D} renderings~\cite{fixmypose} or to describe a single static pose~\cite{posescript}.
Nevertheless, there currently exists no dataset that associates pairs of 3D poses with textual instructions to move from one source pose to one target pose.
In this work, we thus introduce the \dname dataset, which contains over 6,000 textual \textit{modifiers} written by human annotators for this scenario. In addition, we design a pipeline similar to ~\cite{posescript}, to generate modifiers automatically and increase the size of the data, see Figure~\ref{fig:annotation_examples} for some examples.

Leveraging the \dname dataset, we tackle two tasks: \emph{text-based pose editing}, where the goal is to generate new poses from an initial pose and modification instructions, and \emph{correctional text generation} where the objective is to produce a textual description of the difference between a pair of poses (see Figure~\ref{fig:front}). For the first task, we use a baseline consisting in a conditional Variational Auto-Encoder (cVAE). For the second, we consider a baseline built from an auto-regressive transformer model. We provide a detailed evaluation of both baselines, and show promising results.
 
In summary, our contributions are threefold:
\begin{itemize}[noitemsep,topsep=0pt]
    \item[$\circ$] We introduce the \dname dataset (Section~\ref{sec:dataset}) that associates pairs of 3D human poses and human-written textual descriptions of their differences.
    \item[$\circ$] We introduce the task of text-based pose editing (Section~\ref{sec:pose_editing}), that can be tackled with a cVAE baseline.
    \item[$\circ$] We study the task of correctional text generation with a conditioned auto-regressive model (Section~\ref{sec:caption}).
\end{itemize}
\section{Related Work}

\vspace{1mm}
\noindent{\bf 3D pose and text datasets.} AMASS~\cite{amass} gathers several datasets of 3D human motions in SMPL~\cite{smpl} format. BABEL~\cite{babel} and  HumanML3D~\cite{Guo_2022_CVPR} build on top of it to provide free-form text descriptions of the sequences, similarly to the earlier and smaller  Kit Motion-Language dataset~\cite{plappert2016kit}. These datasets focus on sequence semantic (high-level actions) rather than individual pose semantic (fine-grained egocentric relations). To complement, PoseScript~\cite{posescript} links static 3D human poses with descriptions in natural language about fine-grained pose aspects. However, PoseScript does not make it possible to relate two poses together in a straightforward way, as we attempt by introducing the new \dname dataset. In contrast to FixMyPose~\cite{fixmypose}, the \dname dataset we introduce comprehends poses from more diverse sequences and the textual annotations were collected based on actual 3D data and not synthetic 2D image renderings (reduced depth ambiguity).

\vspace{1mm}
\noindent{\bf 3D human pose generation.} Previous works have mainly focused on the generation of pose sequences, conditioning on music~\cite{lee2019dancing, li2021learn},   context~\cite{Corona_2020_CVPR}, past poses~\cite{yuan2020dlow, Zhang_2021_CVPR}, text labels~\cite{chuan2020action2motion, petrovich21actor} and mostly on text descriptions~\cite{Lin2018GeneratingAV, Text2Action, PairedRecurrentAutoencoders, Ahuja2019Language2PoseNL, Ghosh_2021_ICCV, petrovich2022temos, Guo_2022_CVPR, tevet2022motionclip, chuan2022tm2t, kim2022flame}. Some works push it one step further and also attempt to synthesize the mesh appearance~\cite{hong2022avatarclip, youwang2022clipactor}, leveraging large pretrained models like CLIP~\cite{clip}. Similarly to PoseScript~\cite{posescript}, we depart from generic actions and focus on static poses and fine-grained aspects of the human body, to learn about precise egocentric relations. However, we consider two poses instead of one to comprehend detailed pose modifications. Different from ProtoRes~\cite{oreshkin2021protores}, which proposes to manually design a human pose inside a 3D environment based on sparse constraints, we use text for controllability. 
As PoseScript and VPoser~\cite{smplx}, an (unconditioned) pose prior, we use a VAE-based~\cite{vae} model to generate the 3D human poses.

\vspace{1mm}
\noindent{\bf Pose correctional feedback generation.} 
Recent advances in text generation have led to a shift from recurrent neural networks~\cite{sutskever2011generating} to large pretrained transformer models, such as GPT~\cite{brown2020gpt3}. These models can be effectively conditioned using prompting~\cite{ouyang2022training} or cross-attention mechanisms~\cite{radford2022robust-whisper}. While multi-modal text generation tasks, such as image captioning, have been extensively studied~\cite{coco, herdade2019image, vinyals2015show} no previous work has focused on using 3D human poses to generate free-form feedback. In this regard, AIFit~\cite{aifit} extracts 3D data to compare the video performance of a trainee against a coach's and provides feedback based on predefined templates. \cite{zhao20223d} also outputs predefined texts for a small set of exercises and~\cite{liu2022posecoach} does not provide any natural language instructions, either. Besides, FixMyPose~\cite{fixmypose} is based on highly-synthetic 2D images.

\vspace{1mm}
\noindent{\bf Compositional learning} consists in using a query made of multiple distinct elements, which can be of different modalities, as for visual question answering~\cite{antol2015vqa} or composed image retrieval~\cite{vo2019composing}. Similarly to the latter, we are interested in bi-modal queries composed of a textual ``modifier'' which specifies changes to apply on the first element. Modifiers first took the form of single-word attributes~\cite{parikh2011relative, nagarajan2018attributes, doughty2020action} and evolved into free-form texts~\cite{fashioniq, cirplant}. While a large body of works focus on text-conditioned image editing~\cite{talk2edit, instructpix2pix, hertz2022prompt} or text-enhanced image search~\cite{vo2019composing, Baldrati_2022_CVPR, delmas2022artemis}, few study 3D human body poses.
ClipFace~\cite{aneja2022clipface} proposes to edit 3D morphable face models and StyleGAN-Human~\cite{fu2022stylegan} generates 2D images of human bodies in very model-like poses. PoseTutor~\cite{dittakavi2022pose} provides an approach to highlight joints with incorrect angles on 2D yoga/pilate/kung-fu images.
More related to our work, FixMyPose~\cite{fixmypose} performs composed image retrieval. Conversely to them, we propose to \textit{generate} a \textit{3D} pose based on an initial static pose and a modifier expressed in natural language.
\section{The \dname dataset}
\label{sec:dataset}

To tackle the two pose correctional tasks considered in this paper, we introduce the new \dname dataset. It consists of 135k triplets of \{\textit{pose A}, \textit{pose B}, \textit{text modifier}\}, where pose $B$ (the target pose) is the result of the correction of pose $A$ (the source pose), as specified by the \textit{text modifier}. The 3D human body poses were sampled from AMASS~\cite{amass}. All pairs were captioned in Natural Language thanks to our automatic comparative pipeline; 6157 pairs were additionally presented to human annotators on the crowd-source annotation platform Amazon Mechanical Turk.
We next present the pair selection method, the annotations process and some dataset statistics.


\subsection{Pair selection process}

\vspace{0.5mm}
\noindent{\bf In- and Out-of-sequence pairs.}
Pose pairs can be of two types:  ``\textit{in-sequence}" (IS) or ``\textit{out-of-sequence}" (OOS). In the first case, the two poses belong to the same AMASS sequence and are temporally ordered (pose $A$ happens before pose $B$). We select them with a maximum time difference of 0.5 second, to have both textual modifiers describing precisely atomic motion sub-sequences and ground-truth motion. For an increased time difference between the two poses, they could be an infinity of plausible in-between motions, which would weaken such supervision signal. Out-of-sequence pairs are made of two poses from different sequences; to help generalize to less common motions and to study poses of similar configuration but different style, empowering ``pose correction'' beside ``motion continuation''.

\vspace{0.5mm}
\noindent{\bf Selecting pose B.}
As we aim to obtain pose $B$ from pose $A$, we consider that pose $B$ is guiding the most the annotation: while the text modifier should account for pose $A$ and refer to it, its true target is pose $B$. Thus, to build the triplets, we first choose the set of poses $B$. So to maximize the diversity of poses, we follow~\cite{posescript}, and get a set $S$ of 100k poses sampled with a farthest-point algorithm. Poses $B$ are then iteratively selected from $S$.

\vspace{0.5mm}
\noindent{\bf Selecting pose A.}
The paired poses should satisfy two main constraints.
First, poses $A$ and $B$ should be similar enough for the text \textit{modifier} not to become a complete \textit{description} of pose B: if \textit{A} and \textit{B} are too different, it is easier for the annotator to just ignore \textit{A} and directly characterize \textit{B}~\cite{fashioniq, cirplant}. Yet, we aim at learning fine-grained and subtle differences between two poses. Hence, we rank all poses in $S$ with regard to each pose $B$ based on the cosine similarity of their PoseScript semantic pose features~\cite{posescript}. Pose $A$ is to be selected within the top 100. 
Second, the two poses should be different enough, so that the modifier does not collapse to oversimple instructions like `\textit{raise your right hand}', which would not compare to realistic scenarios.
While we expect the poses to be quite different as they belong to $S$, we go one step further and leverage posecode information~\cite{posescript} to ensure that the two poses have at least 15 (resp.\ 20) low-level different properties for IS (resp.\ OOS) pairs.

\vspace{0.5mm}
\noindent{\bf One- and Two-way pairs.}
We consider all possible IS pairs $A \to B$, with $A$ and $B$ in $S$, that meet the selection constraints. Then, following the order defined by $S$, we sample OOS pairs: for each selected pair $A \to B$, if $A$ was not already used for another pair, we also consider $B \to A$. We call such pairs `\textit{two-way}' pairs, as opposed to `\textit{one-way}' pairs. Two-way pairs could be used for cycle consistency.

\begin{figure}[t]
\centering
\includegraphics[width=0.44\linewidth]{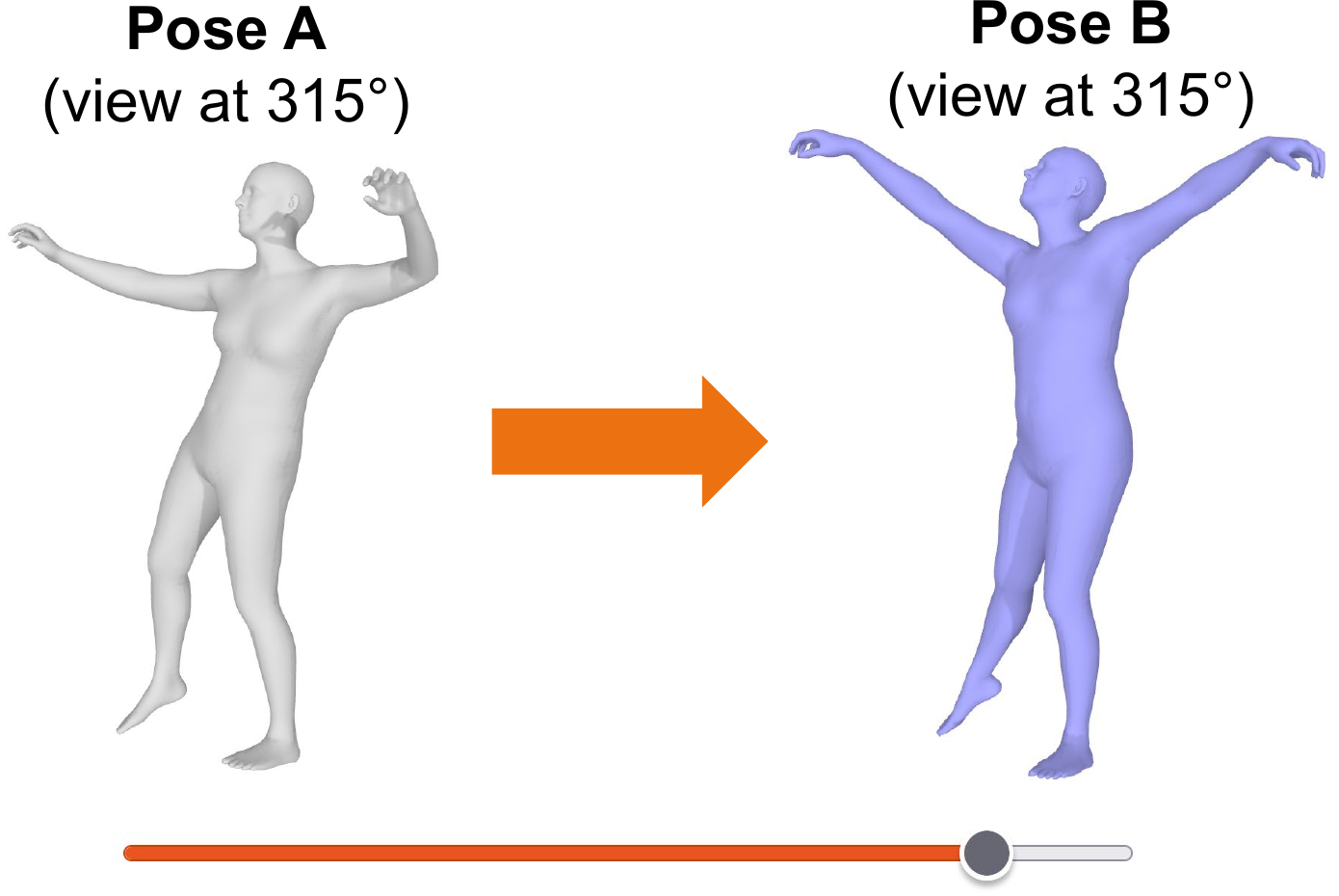} 
\includegraphics[width=0.55\linewidth]{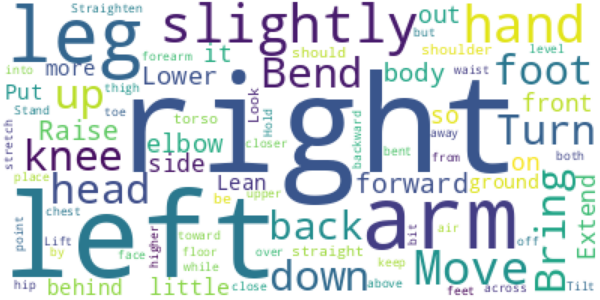} \\[-0.25cm]
\caption{\textbf{Left: Data presented to the annotators.} The slider makes it possible to look at the poses under different viewpoints. \textbf{Right: word cloud} of the \dname annotations.}
\label{fig:annotation_interface}

\vspace{-0.2cm}

\end{figure}
\begin{table}
    \centering
    \resizebox{\linewidth}{!}{
    \begin{tabular}{lrl}
    \toprule
    Property & Proportion & Example \\
    \toprule
    Egocentric relations & 74\% & \textit{Join your hands \textbf{in front of} your chest.} \\[0.05cm]
    Analogies & 5\% & \textit{... \textbf{like you're} about to clap your hands.} \\[0.05cm]
    Implicit side description & 25\% & \textit{Place your left toes on the ground} \\ & & \textit{ and extend your \textbf{$\O$} leg slightly.} \\[-0.1cm]
    \bottomrule
    \end{tabular}
    } \\[-0.3cm]
    \caption{\textbf{Semantic analysis} on 104 sampled human texts.}
    \label{tab:semantic_analysis}
\vspace{-0.35cm}
\end{table}
\vspace{0.5mm}

\noindent{\bf Splits.}
We use the same sequence-based split as~\cite{posescript}, and perform pose pair selection independently in each subset.
Since we also use the same ordered set $S$, some poses are annotated both with a description and a modifier: such complementary information can be used in a multitask setting.


\subsection{Collection of human annotations}

We collected the textual modifiers on Amazon Mechanical Turk from English-speaking annotators with a 95\% approval rate who already completed at least 5000 tasks. To limit perspective-based mistakes, we presented both poses rendered under different viewpoints (see Figure~\ref{fig:annotation_interface}, left). An annotation could not be submitted until it was more than 10 words and several viewpoints were considered. The orientation of the poses was normalized so they would both face the annotator in the front view. Only for in-sequence pairs, we would apply the normalization of pose $A$ to pose $B$, to stay faithful to the global change of orientation in the ground-truth motion sequences.

The annotators were given the following instruction: ``\textit{You are a coach or a trainer. Your student is in pose A, but should be in pose B. Please write the instructions so they can correct the pose on at least 3 aspects.}''. Annotators were required to describe the position of the body parts relatively to the others (\eg `\textit{Your right hand should be close to your neck.}'), to use directions (such as `\textit{left}' and `\textit{right}') in the subject's frame of reference and to mention the rotation of the body, if any. They were also encouraged to use analogies (\eg `\textit{in a push-up pose}'). For the annotations to size-agnostic, distance metrics were prohibited.

The task was first made available to any worker by tiny batches. Annotations were carefully scrutinized, and only the best workers were qualified to pursue to larger batches, with lighter supervision. In total, about 15\% of the annotations were manually reviewed, and corrected when needed. We further cleaned the annotations by fixing misspelled and duplicated words, detected automatically.
Figure~\ref{fig:annotation_examples} shows some pose pairs and their annotated modifiers.


\subsection{Generating annotations automatically}

To scale up the dataset, we design a pipeline to automatically generate thousands of modifiers, by relying on low-level properties as in~\cite{posescript}.
The process takes as input the 3D keypoint positions of two poses $A$ and $B$, and outputs a textual instruction to obtain pose $B$ from pose $A$.
First, it measures and classifies the variation of atomic pose configurations to obtain a set of ``\textit{paircodes}''. For instance, we attend to the motion of the keypoints along each axis (``\textit{move the right hand slightly to the left}'' (x-axis), ``\textit{lift the left knee}'' (y-axis)), to the variation of distance between two keypoints (``\textit{move your hands closer}'') or to the angle change (``\textit{bend your left elbow}'').
We further define ``\textit{super-paircodes}'', resulting from the combination of several paircodes or posecodes~\cite{posescript}; \eg the paircode ``\textit{bend the left knee less}'', associated to the posecode ``\textit{the left knee is slightly bent}'' on pose $A$ leads to the super-paircode ``\textit{straighten the left leg}''. The super-paircodes make it possible to describe higher-level concepts or to refine some assessments (\eg only tell to move the hands farther away from each other if they are close to begin with).
The paircodes are next aggregated using the same set of rules as in~\cite{posescript}, then they are structurally ordered, to gather information about the same general part of the body within the description.
Ultimately, for each paircode, we sample and complete one of the associated template sentences. Their concatenation yields the automatic modifier. 
Please refer to the supplementary for more details.
The whole process produced 135k annotations in less than 15 minutes. Some examples are shown in Figure~\ref{fig:annotation_examples}.
In this paper, we use the automatic data for pretraining only.


\subsection{Statistics and semantic analysis}

\begin{table}
\centering
\begin{minipage}[t]{.49\linewidth}
    \resizebox{\linewidth}{!}{
    \begin{tabular}{lrr}
        \toprule
        & \textit{automatic} & \textit{human} \\
        \midrule
        in-sequence & 25,201 & 2,615 \\
        out-of-sequence & 110,104 & 3,542 \\
        \midrule
        both-way & 93,180 & 2,710 \\
        one-way & 42,125 & 3,447 \\    
        \midrule
        total & 135,305 & 6,157 \\
        \bottomrule
    \end{tabular}
    } \\[-0.2cm]
    \caption{\textbf{Number of pairs} of each set and type.}
    \label{tab:dataset_pair_info}
\end{minipage}
\hfill
\begin{minipage}[t]{.49\linewidth}
    \resizebox{\linewidth}{!}{
    \begin{tabular}{lrr}
        \toprule
        & \textit{automatic} & \textit{human} \\
        \midrule
        different poses  & 99,231 & 7,433 \\
        different poses A & 87,793 & 5,343 \\
        different poses B & 98,939 & 5,922 \\
        \midrule
        ~~~in PoseScript & 6,249 & 3,551 \\
        A in PoseScript & 6,160 & 2,753 \\
        B in PoseScript & 6,226 & 3,143 \\
        \bottomrule
    \end{tabular}
    } \\[-0.2cm]
    \caption{\textbf{Number of poses} per type or shared with ~\cite{posescript}.}
    \label{tab:dataset_pose_info}
\end{minipage}

\vspace{-0.3cm}

\end{table}

\dname contains 6157 (resp.\ 135k) human- (resp.\ automatically-) annotated pairs, split according to a 70\%-10\%-20\% proportion. In average, human-written text modifiers are close to 30 words long with a minimum of 10 words. All together, they form a cleaned vocabulary of 1068 words, a wordcloud of which is shown in Figure~\ref{fig:annotation_interface} (right).

Negation particles were detected in 3.6\% of the annotations, which makes textual queries with negations a bit harder, akin to similar datasets~\cite{fashioniq, posescript}.
A semantic analysis carried out on 104 annotations taken at random is reported in Table~\ref{tab:semantic_analysis}. We found that textual modifiers provide correctional instructions about 4 different body parts in average, which vary depending on the context (pose $A$).

\begin{figure*}
\centering
\includegraphics[width=\textwidth]{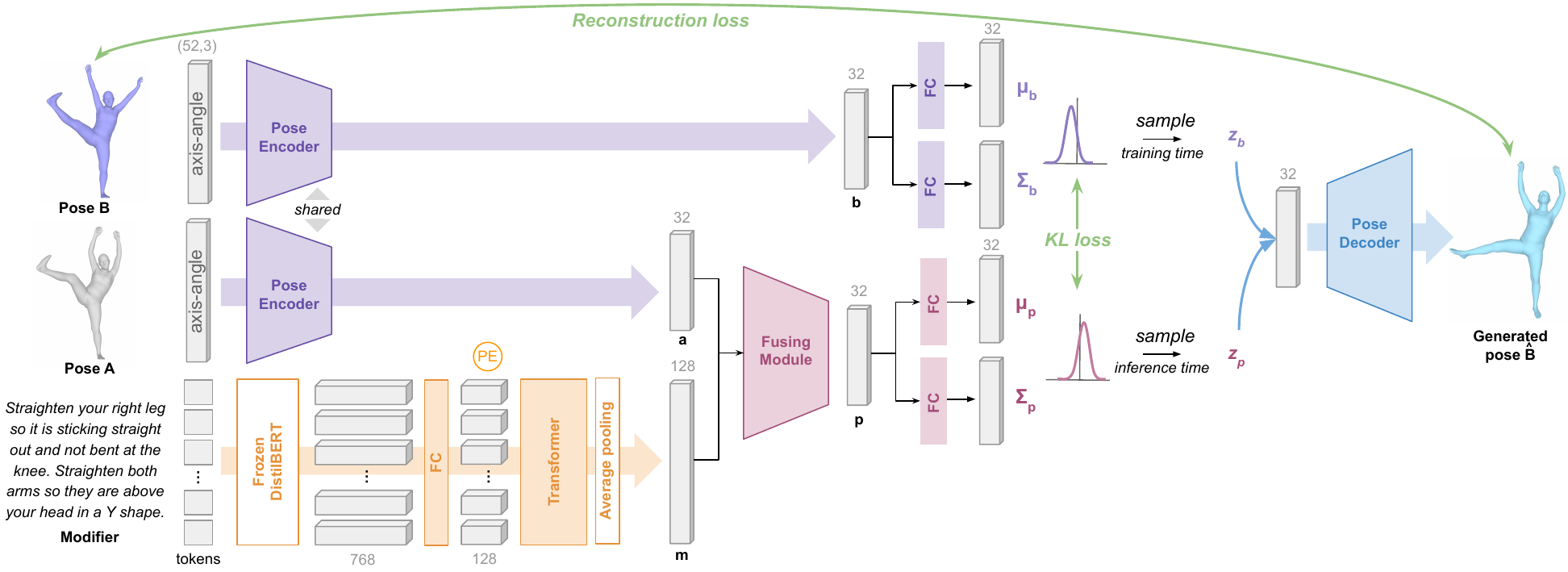} \\[-0.3cm]
\caption{\textbf{Overview of our text-based pose editing baseline.} 
The top part represents a standard VAE, where poses are encoded into a Gaussian distribution. At training time, a latent variable is sampled and decoded into a pose to learn pose reconstruction. The bottom left part represents the conditioning: the text is encoded using a frozen DistilBERT with a small transformer on top. It is combined with source pose features in the \textit{fusion module}, from which we predict a Gaussian distribution. A KL loss ensures the alignment of the distributions from the standard VAE and the conditioning. At test time, we sample from the latter to predict the target pose.}
\label{fig:method_pose_generation}
\end{figure*}

A few other annotation behaviors were found to be quite difficult to quantify, in particular ``missing'' instructions. Sometimes, details are omitted in the text because the context given by pose $A$ is ``taken for granted''. For instance, in the 3rd example shown in Figure~\ref{fig:annotation_examples}, the ``\textit{45-degree angle}'' is to be understood with regard to the ``0 degree'' plan defined by the back of the body in pose $A$. Moreover, the annotator did not specify how the position of the arms have changed, supposedly because this change comes naturally once the back is straighten up, from the structure of the kinematic chain. These challenges are inherent to the task.

Detailed statistics are presented in Tables~\ref{tab:dataset_pair_info} and~\ref{tab:dataset_pose_info}.

\section{Application to Text-based Pose Editing}
\label{sec:pose_editing}

We introduce a VAE~\cite{vae} baseline to perform text-based 3D human pose editing. Specifically, we aim to generate plausible new poses based on two input elements: an initial pose $A$ providing some context (a starting point for modifications), and a textual modifier which specifies the changes to be made. Figure~\ref{fig:method_pose_generation} gives an overview of our model.

\myparagraph{Data processing.} Poses are characterized by their SMPL-H~\cite{mano} body joint rotations in axis-angle representation. Their global orientation is first normalized along the y-axis. For in-sequence pairs, the same normalization that was applied to pose $A$ is applied to pose $B$ in order to preserve information about the change of global orientation.

\myparagraph{Training phase.}
During training, the model encodes both the query pose $A$ and the ground-truth target pose $B$ using a shared pose encoder, yielding respectively features $\ba$ and $\bb$ in $\mathbb{R}^d$.
The tokenized text modifier is fed into a pretrained embedding module to extract expressive word encodings. These are further processed by a learned textual model, to yield a global textual representation $\bm \in \mathbb{R}^n$.
Next, the two input embeddings $\ba$ and $\bm$ are provided to a fusing module which outputs a single vector $\bp \in \mathbb{R}^d$. Both $\bb$ and $\bp$ then go through specific fully connected layers to produce the parameters of two Gaussian distributions: the posterior $\mathcal{N}_b = \mathcal{N}(\cdot |\bmu(\bb), \bSigma(\bb))$ and the prior $\mathcal{N}_p = \mathcal{N}(\cdot | \bmu(\bp), \bSigma(\bp))$ conditioned on $\bp$ from the fusion of $\ba$ and $\bm$. Eventually, a sampled latent variable $\bz_b \sim \mathcal{N}_b$ is decoded into a reconstructed pose $\hat{B}$.

\indent The loss consists in the sum of a reconstruction term $\mathcal{L}_R(B, \hat{B})$ and the Kullback-Leibler (KL) divergence between $\mathcal{N}_b$ and $\mathcal{N}_p$. The former enables the generation of plausible poses, while the latter acts as a regularization term to align the two spaces. The combined loss is then:
\begin{equation}
    \mathcal{L_{\text{pose editing}}} = \mathcal{L}_R(B, \hat{B}) + \mathcal{L}_{KL}(\mathcal{N}_b, \mathcal{N}_p).
\end{equation}

We use the same negative log likelihood-based reconstruction loss as in~\cite{posescript}: it is applied to the output joint rotations in the continuous 6D representation~\cite{zhou2019continuity}, and both the joint and vertices positions inferred from the output by the SMPL-H~\cite{mano} model.

\myparagraph{Inference phase.} The input pose $A$ and the text modifier are processed as in the training phase. However, this time we sample $\bz_p~\sim~\mathcal{N}_p$ to obtain the generated pose $\hat{B}$.

\myparagraph{Evaluation metrics.} We report the Evidence Lower Bound (ELBO) for the size-normalized rotations, joints and vertices, as well as the Fr\'echet inception distance (FID) which compares the distribution of the generated poses with the one of the expected poses, based on their semantic PoseScript features. The ELBO and the FID are mostly sensitive to complementary traits (support coverage and sample quality respectively). When some settings do not improve all metrics; we then base our decisions on the metrics with the highest differential. While the ELBO is better suited to evaluate generative models than reconstruction metrics, for intuitiveness, we also report the the MPJE (mean-per-joint error, in mm), the MPVE (mean-per-vertex error, in mm) and the geodesic distance for joint rotations (in degrees) between the target and the best (\ie, closest) generated sample out of $N{=}30$ in all experiments.

\myparagraph{Architecture details and ablations.} 
We use the VPoser~\cite{smplx} architecture for the pose auto-encoder, resulting in features of dimension $d=32$. The variance of the decoder is considered a learned constant~\cite{sigmavae}. 
We experiment with two different text encoders (Table~\ref{tab:pose_generation_ablatives}, top): (i) a bi-GRU~\cite{bigru} mounted on top of pretrained GloVe word embeddings~\cite{pennington2014glove}, or (ii) a transformer followed by average-pooling, processing frozen DistilBERT~\cite{sanh2019distilbert} word embeddings. We find that the transformer pipeline outperforms the other in terms of ELBO (+0.24 in average) when no additional pretraining is involved, supposedly because it uses already strong general-pretrained weights. Pretraining on our automatic modifiers brings the bi-GRU pipeline on par with the transformer one (+0.04). For simplicity, we will thereafter resort to the former.

For fusion, we use TIRG~\cite{vo2019composing}, a well-spread module for compositional learning. It consists in a gating mechanism composed of two 2-layer Multi-Layer Perceptrons (MLP) $f$ and $g$ balanced by learned scalars $w_f$ and $w_g$ such that:
\begin{equation}
    \bp = w_f f([\ba, \bm]) \odot \ba + w_g g([\ba,\bm]).
\end{equation}
It is designed to `preserve' the main modality feature $\ba$ while applying the modification as a residual connection.

\begin{table}[t!]
    \centering
    \resizebox{\linewidth}{!}{%
    \begin{tabular}{l@{}lccccccc}
    \toprule
    \multirow{2}{*}{} & \multirow{2}{*}{} & \multirow{2}{*}{FID $\downarrow$} & \multicolumn{3}{c}{ELBO $\uparrow$} & \multicolumn{3}{c}{Reconstruction $\downarrow$ \textit{(best of 30)}} \\
    \cmidrule(lr){4-6} \cmidrule(lr){7-9}
    & & & jts & v2v & rot & MPJE & MPVE & Geodesic \\
    \midrule
    \rowcolor{myrowcolor}
    \multicolumn{9}{l}{\textbf{Text Encoder} \textit{(with/without pretraining)}} \\
    \multirow{2}{*}{\textit{without}} & ~~~~ GloVe + bi-GRU & 0.19 & 0.61 & \textbf{1.51} & 0.50 & 278 & 217 & 9.89 \\
    & ~~~~ DistilBERT+transformer & \textbf{0.10} & \textbf{0.95} & \textbf{1.51} & \textbf{0.63} & \textbf{226} & \textbf{180} & \textbf{9.22} \\
    \cmidrule(lr){2-9}
    \multirow{2}{*}{\textit{with}} & ~~~~ GloVe + bi-GRU & \textbf{0.02} & \textbf{1.40} & \textbf{1.88} & \textbf{0.99} & \textbf{199} & \textbf{165} & \textbf{8.59} \\
    & ~~~~ DistilBERT+transformer & \textbf{0.02} & 1.37 & 1.84 & 0.93 & 201 & 167 & 8.70 \\
    \midrule
    \rowcolor{myrowcolor}
    \multicolumn{9}{l}{\textbf{Data augmentations} \textit{(with/without pretraining, GloVe+bi-GRU config)}} \\
    \multirow{6}{*}{\textit{without}} & ~~ no augmentation & 0.19 & 0.61 & \underline{1.51} & 0.50 & 278 & 217 & 9.89 \\
    & ~~ + L/R flip & 0.13 & \textbf{1.10} & \textbf{1.73} & \underline{0.58} & 250 & 196 & 9.57 \\
    & ~~ + paraphrases & 0.19 & 0.90 & 1.45 & \underline{0.58} & \underline{233} & \underline{186} & \underline{9.44} \\
    & ~~ + PoseMix & \underline{0.10} & 0.63 & 1.12 & \underline{0.58} & 254 & 202 & 9.61 \\
    & ~~ + PoseMix \& PoseCopy & \textbf{0.04} & \underline{1.03} & 1.50 &\textbf{ 0.78} & \textbf{221} & \textbf{178} & \textbf{9.07} \\
    \cmidrule(lr){2-9}
    \multirow{6}{*}{\textit{with}} & ~~ no augmentation & \textbf{0.02} & 1.40 & 1.88 & \textbf{0.99} & 199 & 165 & \underline{8.59} \\
    & ~~ + L/R flip & \textbf{0.02} & \textbf{1.47} & \textbf{1.94} & 0.97 & \underline{197} & \underline{163} & 8.65 \\
    & ~~ + paraphrases & \textbf{0.02} & 1.43 & 1.90 & 0.97 & 198 & 164 & \textbf{8.58} \\
    & ~~ + PoseMix & 0.06 & 0.68 & 1.13 & 0.91 & 214 & 174 & 8.74 \\
    & ~~ + PoseMix \& PoseCopy & \underline{0.03} & 1.23 & 1.71 & \underline{0.98} & 208 & 172 & 8.75 \\
    & ~~ + L/R flip \& paraphrases & \textbf{0.02} & \underline{1.44} & \underline{1.92} & 0.97 & \textbf{196} & \textbf{162} & 8.62 \\
    \bottomrule
    \end{tabular}
    } \\[-0.2cm]
\caption{\textbf{Text-based pose editing results} for various architectures, data augmentations and training strategies. We show the best result in bold and underline the second best.}
\label{tab:pose_generation_ablatives}
\vspace{-0.2cm}
\end{table}

\myparagraph{Training data and augmentations ablations.} We experiment with several kinds of data augmentations and training data. Corresponding results are reported in Table~\ref{tab:pose_generation_ablatives} (bottom).
First, we try left/right flipping by swapping the rotations of the left and right body joints (\eg the left hand becomes the right hand) and changing the text accordingly. This improves significantly the relevance of the generated poses (ELBO), especially when the model did not benefit from pretraining on diverse synthetic data (+37\% average improvement of the ELBO).

Next, we use InstructGPT~\cite{ouyang2022training} to obtain 2 paraphrases per annotation. This form of data augmentation was found helpful, particularly when training on a small amount of data, \ie, without pretraining (+20\%).

In order to encourage the model to fully leverage the textual cue, we define \textit{PoseMix}, which gathers both the PoseScript~\cite{posescript} and the \dname datasets. When training with PoseScript data, which consist in pairs of poses and textual descriptions, we set pose $A$ to 0. We notice a mitigated improvement, and even a drop in performance in the pretrained case. One possible reason for that is the difference in formulation between PoseScript descriptions (``\textit{The person is ... with their left hand...}'') and \dname modifiers (``\textit{Move your left hand...}''). Another is that the model then learns to ignore $A$, which is nonetheless crucial in the \dname setting.
To circumvent this last-mentioned issue, we improve the balance of the training data by introducing \textit{PoseCopy}. This consists in providing the model with the same pose in the role of pose $A$ and pose $B$, along with an empty modifier, assuming that a non-existent textual query will force the model to attend pose $A$. The \textit{PoseMix \& PoseCopy} setting yields a great improvement over all metrics for the non-pretrained case (+41\%). This further shows that the formulation gap was not the main issue. As a side product, the fusing branch is now able to work as a pseudo auto-encoder, and to output a copy of the input pose when no modification instruction is provided.

Eventually, the pretraining has a more significant impact than using any kind of data augmentation (+84\%). Besides, the data augmentations become much less effective in this setting (+1\%). The model thus benefits better from pretraining on a large set of new pairs with synthetic instructions, than training on more human-written modifiers of the same pose pairs.
We overall obtain our best model by combining pretraining, left/right flip and paraphrases (last row).

\begin{table}[t!]
    \centering
    \resizebox{\linewidth}{!}{%
    \begin{tabular}{llccccccc}
    \toprule
    \multirow{2}{*}{} & \multirow{2}{*}{FID $\downarrow$} & \multicolumn{3}{c}{ELBO $\uparrow$} & \multicolumn{3}{c}{Reconstruction $\downarrow$ \textit{(best of 30)}} \\
    \cmidrule(lr){3-5} \cmidrule(lr){6-8}
    & & jts & v2v & rot & MPJE & MPVE & Geodesic \\
    \midrule
    \rowcolor{myrowcolor}
    \multicolumn{8}{l}{\textbf{Pair subset}} \\
    ~~in-sequence (530) & 0.04 & 1.33 & 1.78 & 0.88 & \textbf{188} & \textbf{154} & \textbf{8.47} \\
    ~~out-of-sequence (709) & 0.03 & \textbf{1.53} & \textbf{2.02} & \textbf{1.04} & 206 & 168 & 8.80 \\
    ~~full \dname test set (1239) & \textbf{0.02} & 1.44 & 1.92 & 0.97 & 196 & 162 & 8.62 \\
    \midrule
    \rowcolor{myrowcolor}
    \multicolumn{8}{l}{\textbf{Input type} \textit{(full \dname test set - 1239)}} \\
    ~~pose $A$ only & 0.04 & 1.43 & \textbf{1.92} & \textbf{0.97} & 219 & 180 & 8.91 \\
    ~~modifier only & 0.42 & 1.30 & \textbf{1.92} & 0.92 & 378 & 339 & 13.03 \\
    ~~pose $A$ + modifier & \textbf{0.02} & \textbf{1.44} & \textbf{1.92} & \textbf{0.97} & \textbf{196} & \textbf{162} & \textbf{8.62} \\
    \bottomrule
    \end{tabular}
    } \\[-0.2cm]
    \caption{\textbf{Pose editing results for various subsets and input types}, using the best model as per Table~\ref{tab:pose_generation_ablatives}.}
    \label{tab:pose_generation_analysis}
\vspace{-0.3cm}
\end{table}

\begin{figure*}[t!]
\centering
\includegraphics[width=\textwidth]{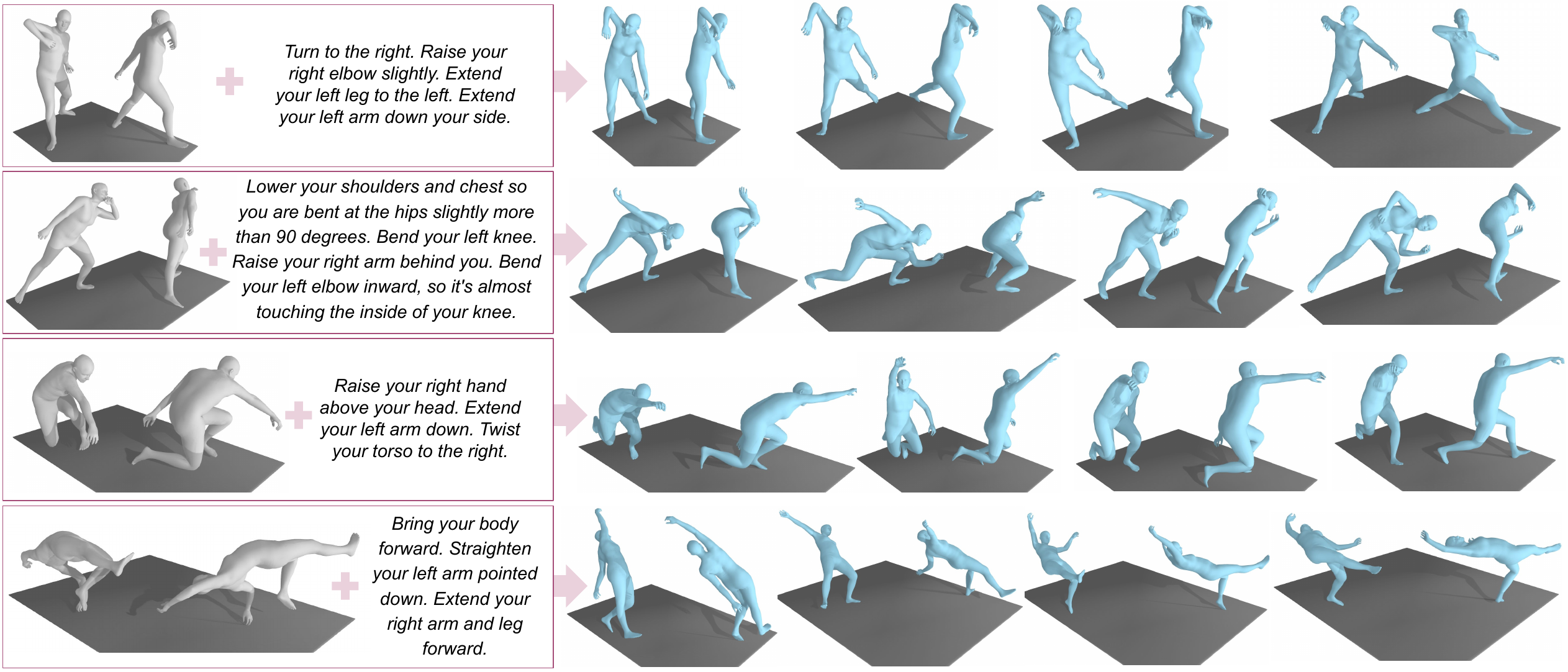} \\[-0.25cm]
\caption{\textbf{Generated poses for the text-based pose editing task} on \dname queries from the left blocks. Two views of each pose are shown on the same ground plane for better visualization of the 3D. Generated poses are shown in blue. Original poses B from the \dname dataset are in the supplementary material.}
\label{fig:b_generation_results}
\vspace{-0.3cm}
\end{figure*}

\myparagraph{Detailed analysis.} In Table~\ref{tab:pose_generation_analysis}, we evaluate our best pose editing model on several subsets of pairs and with different input types. 

First, we notice higher ELBO performance on the out-of-sequence (OOS) pair set compared to the in-sequence (IS) set, suggesting that pairs in the latter are harder. This can be due to pose $A$ and pose $B$ being more similar in IS than OOS, as they belong to the same sequence with a maximum delay of 0.5s. We indeed measure a mean per joint distance of 311mm between $A$ and $B$ in IS \vs 350mm in OOS: the differences between IS poses thus ought to be more subtle, yielding more complex modifiers.
This drop in ELBO performance shows also that the model struggles more with IS modifiers, meaning that it most probably generates, in average, poses that are close to pose $A$, -- in other words, it would takes guesses in the surroundings of pose $A$. This would actually be a good fall-back strategy, because the two poses are rather similar in general. In the IS case, since pose $A$ and pose $B$ are particularly close to each other, the model may end up finding, with enough guesses, a pose closer to pose $B$ than it would in the OOS case, where the two poses are more different. This could explain why the reconstruction metrics using the best sample out of 30 are lower for the IS subset than the OOS subset. 

Next, we compare the results when querying with the \textit{pose $A$ only} or the \textit{modifier only}.
The former achieves already high performance, showing that the initial pose $A$ alone provides a good approximation of the expected pose $B$ -- indeed, the pair selection process constrained pose $A$ and pose $B$ to be quite similar.
The latter yields poor FID and reconstruction metrics: the textual cue is only a modifier, and the same instructions could apply to a large variety of poses. Looking around pose $A$ remains a better strategy than sticking to the sole modifier in order to generate the expected pose. 
Eventually, both parts of the query are complementary: pose $A$ serves as a strong contextual cue, and the modifier guides the search starting from it (the pose being provided through the gating mechanism in TIRG). Both are crucial to reach pose $B$ (last row).

\begin{figure*}[t!]
\centering
\includegraphics[width=\textwidth]{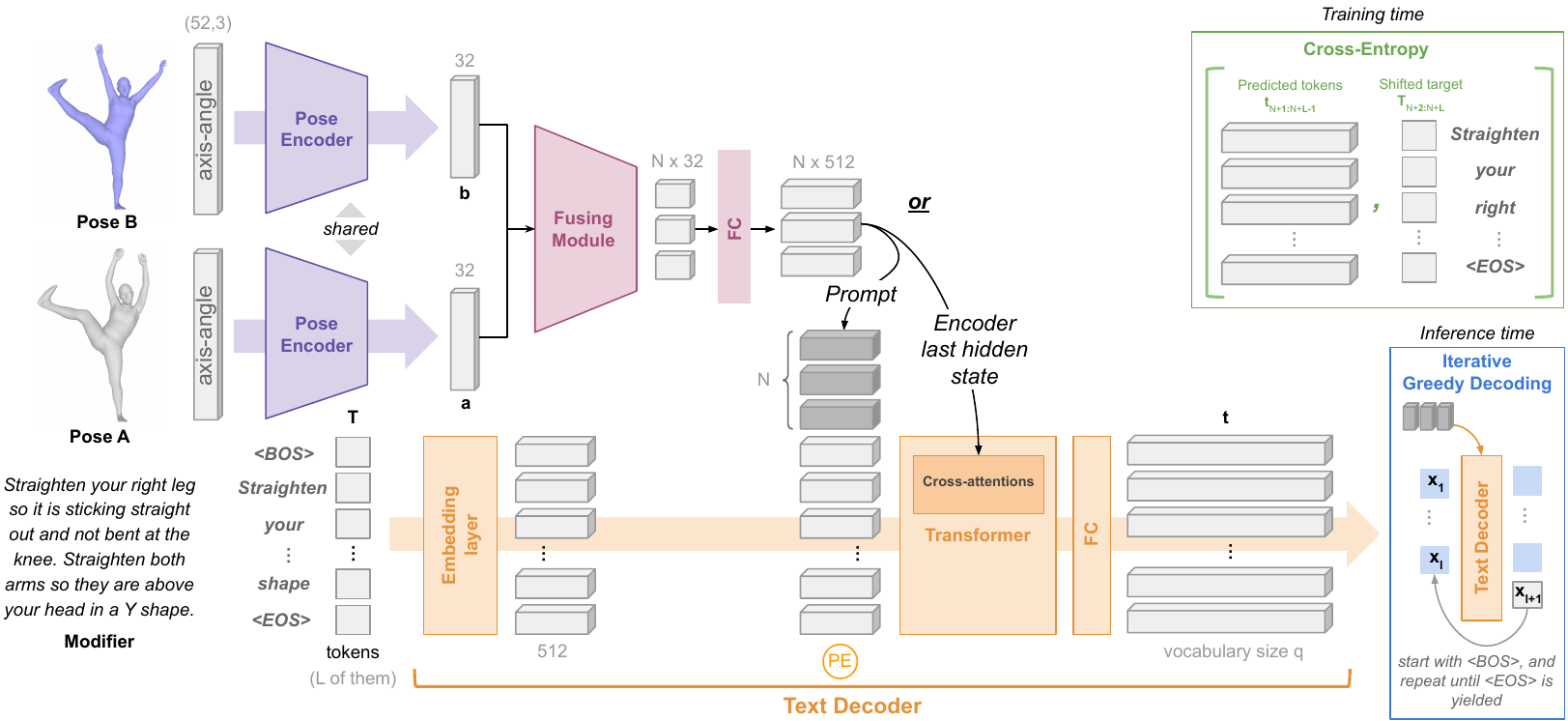} \\[-0.2cm]
\caption{\textbf{Overview of our baseline for correctional text generation.} The bottom part represents a standard auto-regressive transformer model: the next word is predicted from the previously generated tokens. The decoder outputs a distribution of probabilities over the vocabulary for each token.
The top part represents the conditioning on the pose pair: the two pose embeddings are fused together into a set of ``pose tokens'', further used for conditioning via prompting or via cross-attentions in the transformer. At inference, the modifier is generated iteratively using the greedy approach.
}
\label{fig:method_text_generation}
\vspace{-0.3cm}
\end{figure*}

\myparagraph{Qualitative results.} Last, we present qualitative results for text-based 3D human pose editing in Figure~\ref{fig:b_generation_results}. It appears that the model has a relatively good semantic comprehension of the different body parts and of the actions to modify their positions. Some egocentric relations (``\textit{Raise your right elbow slightly.}'', first row) are better understood than others, in particular contact requirements (``\textit{Bend your elbow so it's almost touching the inside of your knee}'', second row). When missing some specifications, the model generates various pose configurations (\eg the extent of the left leg extension in the first example). It can handle a number of instructions at once (third row), but may fail to attend all of them. Crouching and lying-down poses are the most challenging (see failure case in the last row, and how the crouch is hardly preserved in the third row).

\section{Application to correctional text generation}
\label{sec:caption}

We next present a baseline for correctional text generation. We aim to produce feedback in natural language explaining how the source pose $A$ should be modified to obtain the target pose $B$. We rely on an auto-regressive model conditioned on the pose pair, which iteratively predicts the next word given the previous generated ones (see Fig.~\ref{fig:method_text_generation}).

\myparagraph{Training phase.} 
Let $T_{1:L}$ be the $L$ tokens of the text modifier. An auto-regressive generative model seeks to predict the next token $l+1$ from the first $l$ tokens $T_{1:l}$. Let $p(\cdot \vert T_{1:l})$ be the predicted probability distribution over the vocabulary. The model is trained, via a cross-entropy loss, to maximize the probability of generating the ground-truth token $T_{l+1}$ given previous ones: $ p(T_{l+1} \vert T_{1:l})$.

To predict $p(\cdot \vert T_{1:l})$, the tokens $T_{1:l}$ are first embedded, and positional encodings are injected. The result is fed to a series of transformer blocks~\cite{transformer}, and   projected into a space whose dimension is the vocabulary size $q$. Let $\bt \in \mathbb{R}^{q}$ denote the outcome. The probability distribution over the vocabulary for the next token $p( \cdot \vert T_{1:l})$ could be obtained from $Softmax(\bt)$.

Transformer-based auto-regressive models can be trained efficiently using causal attention masks which, for each token $l$, prevent the network from attending all future tokens $l' > l$, in a single pass.

Now, how do poses come into the picture? Pose $A$ and pose $B$ are encoded using a shared encoder, and combined in the fusing module, which outputs a set of $N$ `pose' tokens. To condition the text generation on pose information, we experiment with two alternatives: those pose tokens can either be used for prompting, \ie, added as extra tokens at the beginning of the text modifier, or serve in cross-attention mechanisms within the text transformer.

\myparagraph{Inference phase.} For inference, we provide the model with the pose tokens and the special \textit{${<}BOS{>}$} token which indicates the sequence beginning. We decode the output $\bt_2$ in a greedy fashion, \ie, we predict the next token as the word that maximizes the negative log likelihood. We proceed iteratively, giving previously decoded tokens $T_{1:l}$ to the model so to obtain the subsequent token $l+1$, until the \textit{${<}EOS{>}$} token (denoting the end of the sequence) is decoded.

\begin{table*}[t!]
    \centering
    \resizebox{0.85\textwidth}{!}{%
    \begin{tabular}{llcccccccccc}
        \toprule
        \multirow{2}{*}{} & \multirow{2}{*}{} & \multicolumn{3}{c}{R Precision $\uparrow$} & \multicolumn{3}{c}{NLP $\uparrow$} & \multicolumn{3}{c}{Reconstruction $\downarrow$ \textit{(best of 30)}} \\
        \cmidrule(lr){3-5} \cmidrule(lr){6-8} \cmidrule(lr){9-11}
        & & R@1 & R@2 & R@3 & BLEU-4 & ROUGE-L & METEOR & MPJE & MPVE & Geodesic \\
        \midrule
        \rowcolor{myrowcolor}
        \multicolumn{11}{l}{\textbf{\textit{Control measures}}} \\
        \multicolumn{2}{l}{\textit{random text}} & 3.13 & 6.25 & 9.38 & 7.11 & 26.33 & 26.88 & 225 & 185 & 9.07 \\
        \multicolumn{2}{l}{\textit{original text}} & 62.71 & 74.01 & 79.26 & 100.00 & 100.00 & 100.00 & 196 & 162 & 8.62 \\
        \midrule
        \rowcolor{myrowcolor}
        \multicolumn{11}{l}{\textbf{Injection type} \textit{(with/without pretraining)}} \\
        \multirow{2}{*}{\textit{without}} & ~~~~prompt & 3.63 & 7.10 & 10.73 & 9.74 & \textbf{31.88} & 27.72 & 226 & 184 & 8.94 \\
        & ~~~~cross-attention & \textbf{6.78} & \textbf{12.27} & \textbf{17.35} & \textbf{10.62} & 31.66 & \textbf{28.74} & \textbf{220} & \textbf{180} & \textbf{8.85} \\
        \cmidrule(lr){2-2}
        \multirow{2}{*}{\textit{with}} & ~~~~prompt & 15.09 & 22.28 & 30.35 & 11.15 & 32.58 & 29.76 & 211 & 175 & 8.79 \\
        & ~~~~cross-attention & \textbf{58.43} & \textbf{71.35} & \textbf{77.56} & \textbf{12.19} & \textbf{33.94} & \textbf{31.30} & \textbf{192} & \textbf{161} & \textbf{8.55} \\
        \midrule
        \rowcolor{myrowcolor}
        \multicolumn{11}{l}{\textbf{Data augmentations} \textit{(with pretraining \& cross-attention injection)}} \\
        \multicolumn{2}{l}{~~no augmentation} & 58.43 & 71.35 & 77.56 & \textbf{12.19} & 33.94 & \textbf{31.30} & 192 & 161 & 8.55 \\
        \multicolumn{2}{l}{~~with L/R flip} & \textbf{60.69} & \textbf{71.51} & \textbf{78.85} & 12.14 & \textbf{34.02} & 30.90 & \textbf{189} & \textbf{159} & \textbf{8.54} \\
        \multicolumn{2}{l}{~~with paraphrases} & 53.91 & 67.72 & 74.98 & 10.56 & 33.07 & 30.15 & 194 & 162 & 8.65 \\
        \multicolumn{2}{l}{~~with PoseMix} & 45.12 & 56.66 & 64.89 & 10.94 & 33.22 & 30.12 & 197 & 164 & 8.74 \\
    \bottomrule
    \end{tabular}
    }
    \vspace{-0.3cm}
    \caption{\textbf{Correctional text generation results} for various pose injections and data augmentations. For reference, we also provide numbers for the ground-truth texts and an annotated text chosen at random.}
    \label{tab:text_generation_results}

    \vspace{-0.3cm}
\end{table*}

\begin{figure*}[t!]
\centering
   \includegraphics[width=\textwidth]{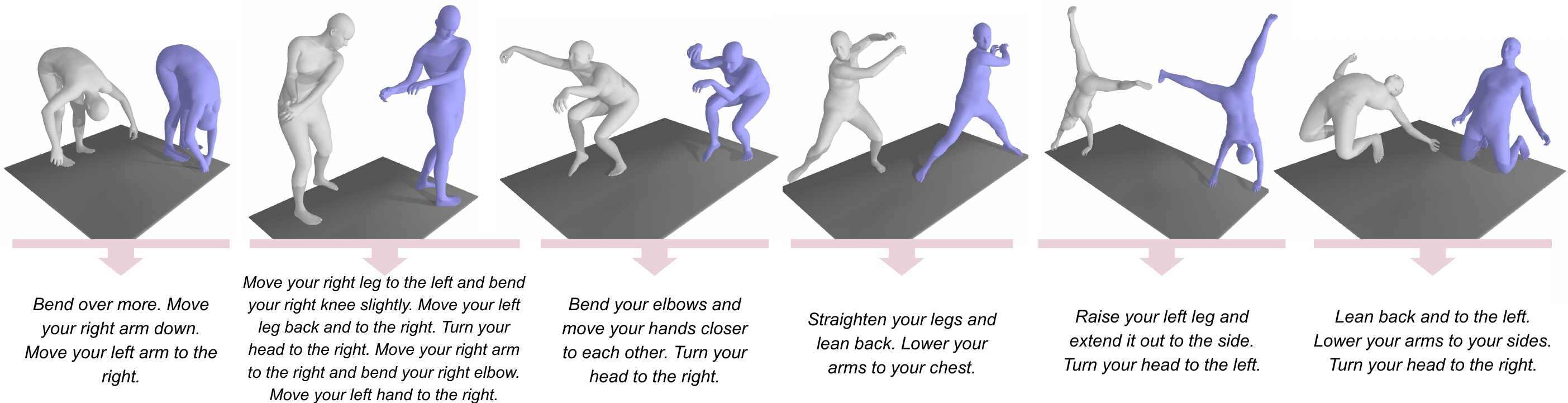} \\[-0.25cm]
\caption{\textbf{Generated correctional texts} for \dname pose pairs (pose $A$ is grey, pose $B$ is purple). The original human annotations for these pose  pairs are available in the supplementary material.
}
\label{fig:text_generation_qualitatives}
\vspace{-0.3cm}
\end{figure*}

\myparagraph{Evaluation metrics.} We resort to standard natural language metrics: BLEU-4~\cite{Papineni02bleu}, Rouge-L~\cite{lin2004rouge} and METEOR~\cite{banarjee2005meteor}, which measure different kinds of n-grams overlaps between the reference text and the generated one. Yet, we notice that these metrics do not reliably reflect the model quality for this task. Indeed, we only have one reference text and, given the initial pose, very different instructions  can lead to the same result (\eg ``\textit{lower your arm at your side}'' and ``\textit{move your right hand next to your hip}''); it is not just a matter of formulation. Thus, we report the top-k R-precision metrics proposed in TM2T~\cite{chuan2022tm2t}: we use contrastive learning to train a joint embedding space for the modifiers and the concatenation of poses $A$ and $B$, then we look at the ranking of the correct pose pair for each generated text within a set of 32 pose pairs. We also report reconstruction metrics on the pose generated thanks to our best model from Section~\ref{sec:pose_editing} using the generated text. These added metrics assess the semantic correctness of the generated texts.

\myparagraph{Quantitative results} are presented in Table~\ref{tab:text_generation_results}. We experiment with the same fusing module as before: TIRG~\cite{vo2019composing}, where the gating applied on the pair leading pose (pose $B$); thus using $N=1$. 
We try prompting and cross-attention to inject the pose information in the text decoder, and found the latter to yield the best results. 
Pretraining on automatic modifiers significantly boosts the performance, \eg with cross-attention injection, the R@2 increases from 12.27\% to 71.35\%.
Regarding data augmentations, the left/right flip yields additional gains (+1.7\% of average R Precision) with results close to those obtained with the ground-truth texts, both for R-precision and reconstruction.
Even if the generated text does not have the same wording as the original text (low NLP metrics), combined with pose A, it achieves to produce a satisfactory pose $\hat{B}$, meaning that it carries the right correctional information. Of course, one should recall that the added metrics rely on imperfect models, which have their own limitations. Finally, we observe a decrease in performance with the paraphrases or the PoseMix settings: we hypothesize that these settings are harder than the regular one for this task, due to new words and formulations.

\myparagraph{Qualitative results.} Fig.~\ref{fig:text_generation_qualitatives} shows some generated texts. The model is able to produce satisfying feedback, it generates egocentric relations (third and fourth examples) and groups indications by body part (second column). However, it tends to mix up pose $A$ and $B$ (last two examples). It also sometimes describes only a subset of the differences.

\section{Conclusion}
\vspace{-1mm}
This paper lays the groundwork for investigating the challenge of correcting 3D human poses using natural language instructions. Going beyond existing methods that utilize language to model global motion or entire body poses, we aim to capture the subtle differences between pairs of body poses, which requires a new level of semantic understanding. For this purpose, we have introduced \dname, a novel dataset with paired poses and their corresponding correctional descriptions. We also presented promising results for two baselines which address the deriving tasks of text-based pose editing and correctional text generation.

\small{\myparagraph{Acknowledgments.} This work is supported by the Spanish government with the project MoHuCo PID2020-120049RB-I00, and by NAVER LABS Europe under technology transfer contract ‘Text4Pose’.}


\clearpage
\bibliographystyle{ieee_fullname}
\bibliography{biblio}

\begin{thebibliography}{10}\itemsep=-1pt

\bibitem{Text2Action}
Hyemin Ahn, Timothy Ha, Yunho Choi, Hwiyeon Yoo, and Songhwai Oh.
\newblock Text2action: Generative adversarial synthesis from language to
  action.
\newblock In {\em ICRA}, 2018.

\bibitem{Ahuja2019Language2PoseNL}
Chaitanya Ahuja and Louis-Philippe Morency.
\newblock Language2pose: Natural language grounded pose forecasting.
\newblock {\em 3DV}, 2019.

\bibitem{aneja2022clipface}
Shivangi Aneja, Justus Thies, Angela Dai, and Matthias Nie{\ss}ner.
\newblock Clipface: Text-guided editing of textured 3d morphable models.
\newblock In {\em SIGGRAPH}, 2023.

\bibitem{antol2015vqa}
Stanislaw Antol, Aishwarya Agrawal, Jiasen Lu, Margaret Mitchell, Dhruv Batra,
  C~Lawrence Zitnick, and Devi Parikh.
\newblock Vqa: Visual question answering.
\newblock In {\em ICCV}, 2015.

\bibitem{Baldrati_2022_CVPR}
Alberto Baldrati, Marco Bertini, Tiberio Uricchio, and Alberto Del~Bimbo.
\newblock Conditioned and composed image retrieval combining and partially
  fine-tuning clip-based features.
\newblock In {\em CVPRW}, 2022.

\bibitem{banarjee2005meteor}
Satanjeev Banerjee and Alon Lavie.
\newblock {METEOR}: An automatic metric for {MT} evaluation with improved
  correlation with human judgments.
\newblock In {\em {ACL} Workshop on Intrinsic and Extrinsic Evaluation Measures
  for Machine Translation and/or Summarization}, 2005.

\bibitem{instructpix2pix}
Tim Brooks, Aleksander Holynski, and Alexei~A Efros.
\newblock Instructpix2pix: Learning to follow image editing instructions.
\newblock In {\em CVPR}, 2023.

\bibitem{brown2020gpt3}
Tom Brown, Benjamin Mann, Nick Ryder, Melanie Subbiah, Jared~D Kaplan, Prafulla
  Dhariwal, Arvind Neelakantan, Pranav Shyam, Girish Sastry, Amanda Askell,
  et~al.
\newblock Language models are few-shot learners.
\newblock In {\em NeurIPS}, 2020.

\bibitem{bigru}
Kyunghyun Cho, Bart Van~Merri{\"e}nboer, Caglar Gulcehre, Dzmitry Bahdanau,
  Fethi Bougares, Holger Schwenk, and Yoshua Bengio.
\newblock Learning phrase representations using rnn encoder-decoder for
  statistical machine translation.
\newblock In {\em EMNLP}, 2014.

\bibitem{Corona_2020_CVPR}
Enric Corona, Albert Pumarola, Guillem Alenya, and Francesc Moreno-Noguer.
\newblock Context-aware human motion prediction.
\newblock In {\em CVPR}, 2020.

\bibitem{delmas2022artemis}
Ginger Delmas, Rafael~Sampaio de Rezende, Gabriela Csurka, and Diane Larlus.
\newblock Artemis: Attention-based retrieval with text-explicit matching and
  implicit similarity.
\newblock In {\em ICLR}, 2022.

\bibitem{posescript}
{Delmas, Ginger and Weinzaepfel, Philippe and Lucas, Thomas and Moreno-Noguer,
  Francesc and Rogez, Gr\'egory}.
\newblock {PoseScript: 3D Human Poses from Natural Language}.
\newblock In {\em {ECCV}}, 2022.

\bibitem{dittakavi2022pose}
Bhat Dittakavi, Divyagna Bavikadi, Sai~Vikas Desai, Soumi Chakraborty, Nishant
  Reddy, Vineeth~N Balasubramanian, Bharathi Callepalli, and Ayon Sharma.
\newblock Pose tutor: An explainable system for pose correction in the wild.
\newblock In {\em CVPR}, 2022.

\bibitem{doughty2020action}
Hazel Doughty, Ivan Laptev, Walterio Mayol-Cuevas, and Dima Damen.
\newblock {A}ction {M}odifiers: {L}earning from {A}dverbs in {I}nstructional
  {V}ideos.
\newblock In {\em CVPR}, 2020.

\bibitem{aifit}
Mihai Fieraru, Mihai Zanfir, Silviu~Cristian Pirlea, Vlad Olaru, and Cristian
  Sminchisescu.
\newblock {AIFit}: Automatic {3D} human-interpretable feedback models for
  fitness training.
\newblock In {\em CVPR}, 2021.

\bibitem{fu2022stylegan}
Jianglin Fu, Shikai Li, Yuming Jiang, Kwan-Yee Lin, Chen Qian, Chen~Change Loy,
  Wayne Wu, and Ziwei Liu.
\newblock Stylegan-human: A data-centric odyssey of human generation.
\newblock In {\em ECCV}, 2022.

\bibitem{Ghosh_2021_ICCV}
Anindita Ghosh, Noshaba Cheema, Cennet Oguz, Christian Theobalt, and Philipp
  Slusallek.
\newblock Synthesis of compositional animations from textual descriptions.
\newblock In {\em ICCV}, 2021.

\bibitem{Guo_2022_CVPR}
Chuan Guo, Shihao Zou, Xinxin Zuo, Sen Wang, Wei Ji, Xingyu Li, and Li Cheng.
\newblock Generating diverse and natural 3d human motions from text.
\newblock In {\em CVPR}, 2022.

\bibitem{chuan2022tm2t}
Chuan Guo, Xinxin Zuo, Sen Wang, and Li Cheng.
\newblock Tm2t: Stochastic and tokenized modeling for the reciprocal generation
  of 3d human motions and texts.
\newblock In {\em ECCV}, 2022.

\bibitem{chuan2020action2motion}
Chuan Guo, Xinxin Zuo, Sen Wang, Shihao Zou, Qingyao Sun, Annan Deng, Minglun
  Gong, and Li Cheng.
\newblock Action2motion: Conditioned generation of {3D} human motions.
\newblock In {\em ACMMM}, 2020.

\bibitem{hendrycks2016gaussian-gelu}
Dan Hendrycks and Kevin Gimpel.
\newblock Gaussian error linear units (gelus).
\newblock {\em arXiv preprint arXiv:1606.08415}, 2016.

\bibitem{herdade2019image}
Simao Herdade, Armin Kappeler, Kofi Boakye, and Joao Soares.
\newblock Image captioning: Transforming objects into words.
\newblock In {\em NeurIPS}, 2019.

\bibitem{hertz2022prompt}
Amir Hertz, Ron Mokady, Jay Tenenbaum, Kfir Aberman, Yael Pritch, and Daniel
  Cohen-Or.
\newblock Prompt-to-prompt image editing with cross attention control.
\newblock In {\em ICLR}, 2023.

\bibitem{hong2022avatarclip}
Fangzhou Hong, Mingyuan Zhang, Liang Pan, Zhongang Cai, Lei Yang, and Ziwei
  Liu.
\newblock Avatarclip: Zero-shot text-driven generation and animation of 3d
  avatars.
\newblock {\em ACM TOG}, 2022.

\bibitem{talk2edit}
Yuming Jiang, Ziqi Huang, Xingang Pan, Chen~Change Loy, and Ziwei Liu.
\newblock Talk-to-edit: Fine-grained facial editing via dialog.
\newblock In {\em ICCV}, 2021.

\bibitem{fixmypose}
Hyounghun Kim, Abhay Zala, Graham Burri, and Mohit Bansal.
\newblock {FixMyPose}: Pose correctional captioning and retrieval.
\newblock In {\em AAAI}, 2021.

\bibitem{kim2022flame}
Jihoon Kim, Jiseob Kim, and Sungjoon Choi.
\newblock Flame: Free-form language-based motion synthesis \& editing.
\newblock In {\em AAAI}, 2023.

\bibitem{kingma2014adam}
Diederik~P Kingma and Jimmy Ba.
\newblock Adam: A method for stochastic optimization.
\newblock In {\em ICLR}, 2015.

\bibitem{vae}
Diederik~P Kingma and Max Welling.
\newblock Auto-encoding variational bayes.
\newblock In {\em ICLR}, 2014.

\bibitem{lee2019dancing}
Hsin-Ying Lee, Xiaodong Yang, Ming-Yu Liu, Ting-Chun Wang, Yu-Ding Lu,
  Ming-Hsuan Yang, and Jan Kautz.
\newblock Dancing to music.
\newblock In {\em NeurIPS}, 2019.

\bibitem{li2021learn}
Ruilong Li, Shan Yang, David~A. Ross, and Angjoo Kanazawa.
\newblock Ai choreographer: Music conditioned 3d dance generation with aist++.
\newblock In {\em ICCV}, 2021.

\bibitem{Lin2018GeneratingAV}
Angela~S. Lin, Lemeng Wu, Rodolfo Corona, Kevin W.~H. Tai, Qixing Huang, and
  Raymond~J. Mooney.
\newblock Generating animated videos of human activities from natural language
  descriptions.
\newblock In {\em NeurIPS workshops}, 2018.

\bibitem{lin2004rouge}
Chin-Yew Lin.
\newblock Rouge: A package for automatic evaluation of summaries.
\newblock In {\em Text summarization branches out}, 2004.

\bibitem{coco}
Tsung-Yi Lin, Michael Maire, Serge Belongie, James Hays, Pietro Perona, Deva
  Ramanan, Piotr Doll{\'a}r, and C~Lawrence Zitnick.
\newblock Microsoft coco: Common objects in context.
\newblock In {\em ECCV}, 2014.

\bibitem{liu2022posecoach}
Jingyuan Liu, Nazmus Saquib, Zhutian Chen, Rubaiat~Habib Kazi, Li-Yi Wei,
  Hongbo Fu, and Chiew-Lan Tai.
\newblock Posecoach: A customizable analysis and visualization system for
  video-based running coaching.
\newblock {\em IEEE trans. VCG}, 2022.

\bibitem{cirplant}
Zheyuan Liu, Cristian Rodriguez-Opazo, Damien Teney, and Stephen Gould.
\newblock Image retrieval on real-life images with pre-trained
  vision-and-language models.
\newblock In {\em ICCV}, 2021.

\bibitem{smpl}
Matthew Loper, Naureen Mahmood, Javier Romero, Gerard Pons-Moll, and Michael~J
  Black.
\newblock {SMPL}: A skinned multi-person linear model.
\newblock {\em ACM TOG}, 2015.

\bibitem{amass}
Naureen Mahmood, Nima Ghorbani, Nikolaus~F Troje, Gerard Pons-Moll, and
  Michael~J Black.
\newblock {AMASS}: Archive of motion capture as surface shapes.
\newblock In {\em ICCV}, 2019.

\bibitem{nagarajan2018attributes}
Tushar Nagarajan and Kristen Grauman.
\newblock Attributes as operators: factorizing unseen attribute-object
  compositions.
\newblock In {\em ECCV}, 2018.

\bibitem{oreshkin2021protores}
Boris~N Oreshkin, Florent Bocquelet, Felix~G Harvey, Bay Raitt, and Dominic
  Laflamme.
\newblock Protores: Proto-residual network for pose authoring via learned
  inverse kinematics.
\newblock In {\em ICLR}, 2021.

\bibitem{ouyang2022training}
Long Ouyang, Jeff Wu, Xu Jiang, Diogo Almeida, Carroll~L Wainwright, Pamela
  Mishkin, Chong Zhang, Sandhini Agarwal, Katarina Slama, Alex Ray, et~al.
\newblock Training language models to follow instructions with human feedback.
\newblock {\em arXiv preprint arXiv:2203.02155}, 2022.

\bibitem{Papineni02bleu}
Kishore Papineni, Salim Roukos, Todd Ward, and Wei jing Zhu.
\newblock Bleu: a method for automatic evaluation of machine translation.
\newblock In {\em ACL}, 2002.

\bibitem{parikh2011relative}
Devi Parikh and Kristen Grauman.
\newblock Relative attributes.
\newblock In {\em ICCV}, 2011.

\bibitem{smplx}
Georgios Pavlakos, Vasileios Choutas, Nima Ghorbani, Timo Bolkart, Ahmed~AA
  Osman, Dimitrios Tzionas, and Michael~J Black.
\newblock Expressive body capture: {3D} hands, face, and body from a single
  image.
\newblock In {\em CVPR}, 2019.

\bibitem{pennington2014glove}
Jeffrey Pennington, Richard Socher, and Christopher~D Manning.
\newblock Glove: Global vectors for word representation.
\newblock In {\em EMNLP}, 2014.

\bibitem{petrovich21actor}
Mathis Petrovich, Michael~J. Black, and G{\"u}l Varol.
\newblock Action-conditioned 3{D} human motion synthesis with transformer
  {VAE}.
\newblock In {\em ICCV}, 2021.

\bibitem{petrovich2022temos}
Mathis Petrovich, Michael~J Black, and G{\"u}l Varol.
\newblock Temos: Generating diverse human motions from textual descriptions.
\newblock In {\em ECCV}, 2022.

\bibitem{plappert2016kit}
Matthias Plappert, Christian Mandery, and Tamim Asfour.
\newblock The kit motion-language dataset.
\newblock {\em Big data}, 2016.

\bibitem{babel}
Abhinanda~R Punnakkal, Arjun Chandrasekaran, Nikos Athanasiou, Alejandra
  Quiros-Ramirez, and Michael~J Black.
\newblock {BABEL}: Bodies, action and behavior with english labels.
\newblock In {\em CVPR}, 2021.

\bibitem{clip}
Alec Radford, Jong~Wook Kim, Chris Hallacy, Aditya Ramesh, Gabriel Goh,
  Sandhini Agarwal, Girish Sastry, Amanda Askell, Pamela Mishkin, Jack Clark,
  et~al.
\newblock Learning transferable visual models from natural language
  supervision.
\newblock In {\em ICML}, 2021.

\bibitem{radford2022robust-whisper}
Alec Radford, Jong~Wook Kim, Tao Xu, Greg Brockman, Christine McLeavey, and
  Ilya Sutskever.
\newblock Robust speech recognition via large-scale weak supervision.
\newblock {\em arXiv preprint arXiv:2212.04356}, 2022.

\bibitem{mano}
Javier Romero, Dimitrios Tzionas, and Michael~J. Black.
\newblock Embodied hands: Modeling and capturing hands and bodies together.
\newblock In {\em SIGGRAPH Asia}, 2017.

\bibitem{sigmavae}
Oleh Rybkin, Kostas Daniilidis, and Sergey Levine.
\newblock Simple and effective vae training with calibrated decoders.
\newblock In {\em ICML}, 2021.

\bibitem{sanh2019distilbert}
Victor Sanh, Lysandre Debut, Julien Chaumond, and Thomas Wolf.
\newblock Distilbert, a distilled version of bert: smaller, faster, cheaper and
  lighter.
\newblock {\em arXiv preprint arXiv:1910.01108}, 2019.

\bibitem{sutskever2011generating}
Ilya Sutskever, James Martens, and Geoffrey~E Hinton.
\newblock Generating text with recurrent neural networks.
\newblock In {\em ICML}, 2011.

\bibitem{tevet2022motionclip}
Guy Tevet, Brian Gordon, Amir Hertz, Amit~H Bermano, and Daniel Cohen-Or.
\newblock Motionclip: Exposing human motion generation to clip space.
\newblock In {\em ECCV}, 2022.

\bibitem{transformer}
Ashish Vaswani, Noam Shazeer, Niki Parmar, Jakob Uszkoreit, Llion Jones,
  Aidan~N Gomez, {\L}ukasz Kaiser, and Illia Polosukhin.
\newblock Attention is all you need.
\newblock In {\em NeurIPS}, 2017.

\bibitem{vinyals2015show}
Oriol Vinyals, Alexander Toshev, Samy Bengio, and Dumitru Erhan.
\newblock Show and tell: A neural image caption generator.
\newblock In {\em CVPR}, 2015.

\bibitem{vo2019composing}
Nam Vo, Lu Jiang, Chen Sun, Kevin Murphy, Li-Jia Li, Li Fei-Fei, and James
  Hays.
\newblock Composing text and image for image retrieval-an empirical odyssey.
\newblock In {\em CVPR}, 2019.

\bibitem{fashioniq}
Hui Wu, Yupeng Gao, Xiaoxiao Guo, Ziad Al-Halah, Steven Rennie, Kristen
  Grauman, and Rogerio Feris.
\newblock Fashion iq: A new dataset towards retrieving images by natural
  language feedback.
\newblock In {\em CVPR}, 2021.

\bibitem{PairedRecurrentAutoencoders}
Tatsuro Yamada, Hiroyuki Matsunaga, and Tetsuya Ogata.
\newblock Paired recurrent autoencoders for bidirectional translation between
  robot actions and linguistic descriptions.
\newblock {\em IEEE RAL}, 2018.

\bibitem{youwang2022clipactor}
Kim Youwang, Kim Ji-Yeon, and Tae-Hyun Oh.
\newblock Clip-actor: Text-driven recommendation and stylization for animating
  human meshes.
\newblock In {\em ECCV}, 2022.

\bibitem{yuan2020dlow}
Ye Yuan and Kris Kitani.
\newblock Dlow: Diversifying latent flows for diverse human motion prediction.
\newblock In {\em ECCV}, 2020.

\bibitem{Zhang_2021_CVPR}
Yan Zhang, Michael~J. Black, and Siyu Tang.
\newblock We are more than our joints: Predicting how 3d bodies move.
\newblock In {\em CVPR}, 2021.

\bibitem{zhao20223d}
Ziyi Zhao, Sena Kiciroglu, Hugues Vinzant, Yuan Cheng, Isinsu Katircioglu,
  Mathieu Salzmann, and Pascal Fua.
\newblock 3d pose based feedback for physical exercises.
\newblock In {\em ACCV}, 2022.

\bibitem{Zhou_2019_CVPR}
Xingran Zhou, Siyu Huang, Bin Li, Yingming Li, Jiachen Li, and Zhongfei Zhang.
\newblock Text guided person image synthesis.
\newblock In {\em CVPR}, 2019.

\bibitem{zhou2019continuity}
Yi Zhou, Connelly Barnes, Jingwan Lu, Jimei Yang, and Hao Li.
\newblock On the continuity of rotation representations in neural networks.
\newblock In {\em CVPR}, 2019.

\end{thebibliography}


\newpage
\appendix
\section*{Supplementary Material}
\setcounter{table}{0}
\renewcommand{\thetable}{A\arabic{table}}
\setcounter{figure}{0}
\renewcommand{\thefigure}{A\arabic{figure}}
\setcounter{equation}{0}
\renewcommand{\theequation}{A\arabic{equation}}
In this supplementary material, we first provide additional details and statistics on the \dname dataset in Section~\ref{app:dataset}.
The original triplets from \dname for the generated results presented in the main paper are available in Section~\ref{app:ground_truth}.
Additional visualizations are provided in Section~\ref{app:misc_visu}.
Finally, we give implementation details in Section~\ref{app:details}.

\section{\dname complementary information}
\label{app:dataset}

In this section, we provide additional details about the creation of the \dname dataset.

\subsection{Human annotations}

\myparagraph{Sequences of origin.} The poses in \dname were extracted from AMASS~\cite{amass}. In Figure~\ref{fig:pose_origin}, we present the proportion of poses coming from each of the datasets included in AMASS. We notice that most poses belong to  the DanceDB dataset (44\%), presumably because this is where the poses are the most diverse. Recall that poses were chosen following a farther-point sampling algorithm to ensure we would get a various subset of poses. Besides, we note that most of the sequences available in DanceDB (94\%) and MPI-limits (83\%) provided at least one pose to \dname, which suggests that \dname could help in apprehending very complex, extreme poses.

\begin{figure}[h]
\begin{center}
   \includegraphics[width=\linewidth]{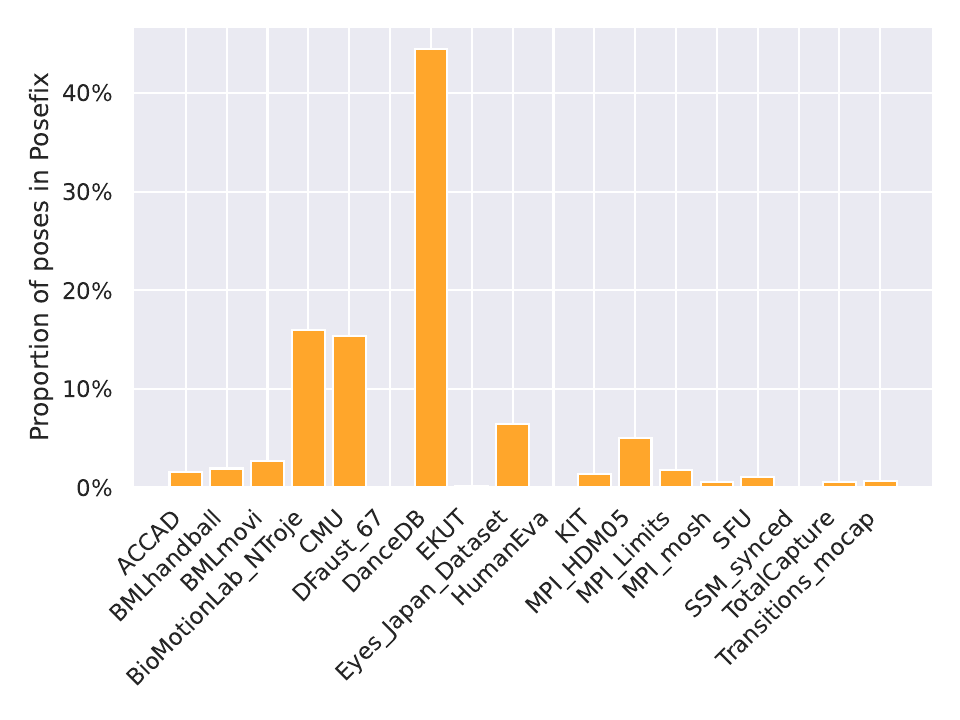} \\[-0.4cm]
  \includegraphics[width=\linewidth]{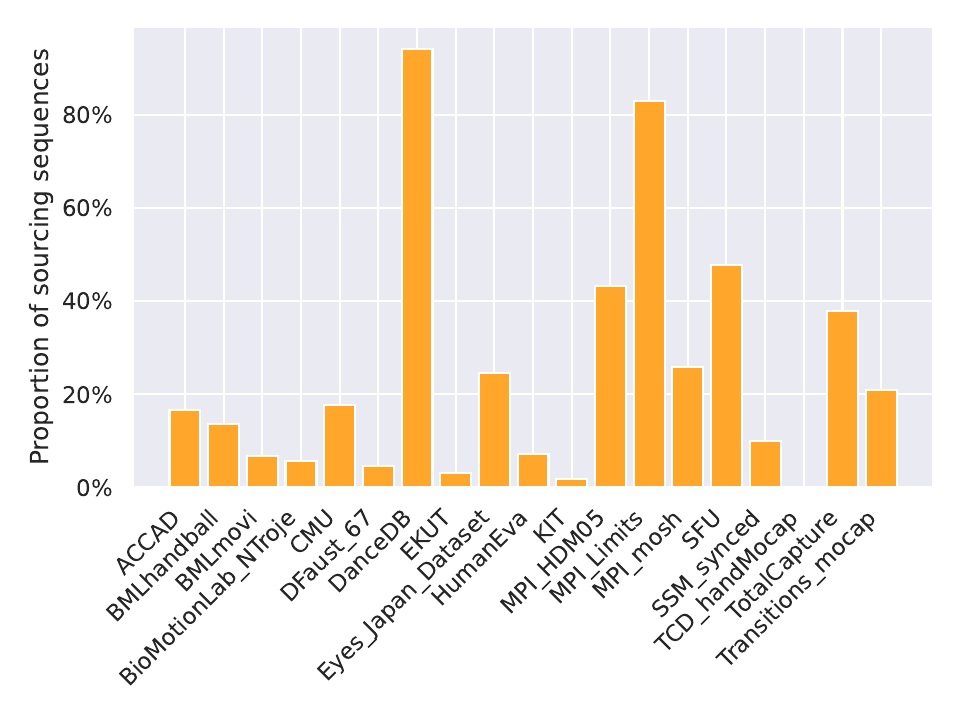} 
\end{center} 
\vspace{-0.5cm}
\caption{\textbf{Origin of the human-annotated poses in \dname.} The top plot shows the proportion of poses in \dname that come from each sub-dataset in AMASS~\cite{amass}. The lower plot shows the proportion of sequences, in each of the sub-dataset, that provided at least one pose to \dname.}
\label{fig:pose_origin}
\end{figure}

\begin{figure}[h]
\begin{center}
   \includegraphics[width=0.85\linewidth]{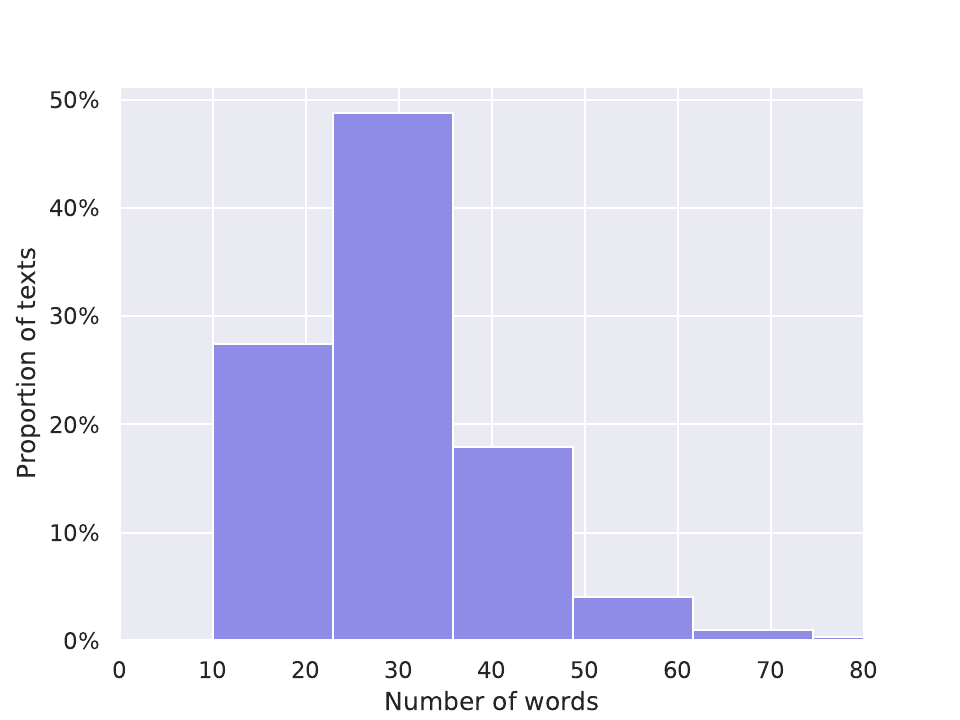}
\end{center}
\vspace{-0.3cm}
\caption{\textbf{Distribution of the number of words in the human-written annotations from \dname}.}
\label{fig:length_hist}
\end{figure}

\myparagraph{Turkers qualifications and statistics.} The annotations were collected on Amazon Mechanical Turk. Participating workers (``Turkers'') had to come from English-speaking countries (Australia, Canada, New Zealand, United Kingdom, USA), have completed at least 5,000 other tasks, and have an approval rate greater than 95\%. In total, 105 different annotators participated. We qualified 20 of them for access to the larger batches, on the basis of at least 3 good annotations. Other 50 workers were excluded from our annotation task because of poor writing, misunderstanding of the task or cheating. The remaining participants did not complete enough annotations of good quality to be qualified for accessing more. Eventually, over 90\% of the annotations were made by 8 annotators.

\myparagraph{Pricing.} Properly completing an annotation, after a bit of training, was timed to take approximately 1'10''. Annotations from the smaller qualifying batches were rewarded \$0.25. Once a worker completed 3 of them correctly, s/he was granted access to the larger batches, where annotations were rewarded \$0.32 each, based on the minimum wage in California for 2023. We additionally paid a 10\% bonus for every 30 annotations.

\begin{figure*}[t!]
\centering
\includegraphics[width=\linewidth]{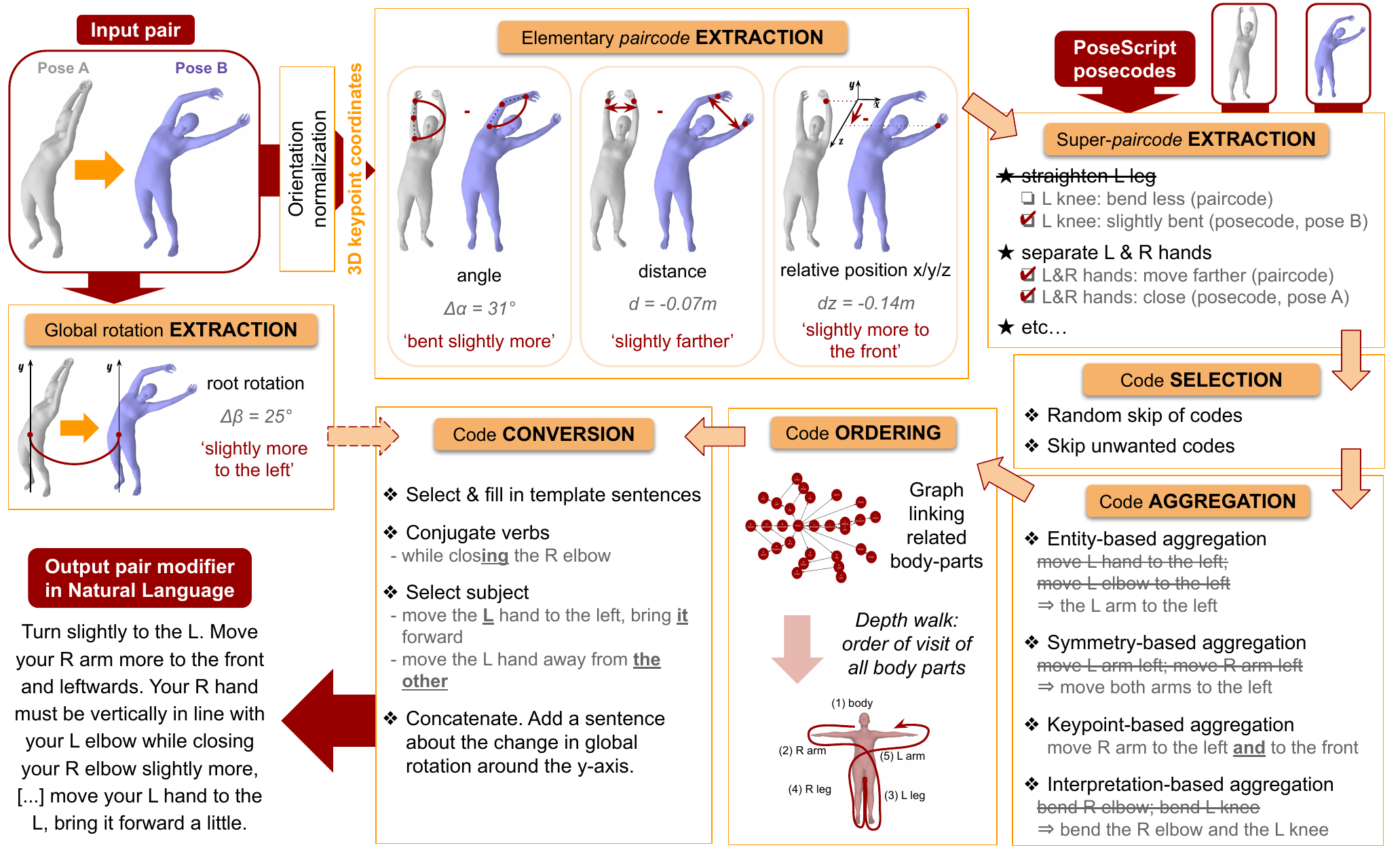} \\[-0.15cm]
\caption{\textbf{Automatic Comparative Pipeline}, which generates modifiers based on the 3D keypoint coordinates of two input poses. L (resp.\ R) stands for `left' (resp.\ `right').}
\label{fig:auto_comparative_pipeline}
\end{figure*}

\myparagraph{Quality assessment.}
Annotations from the early smaller qualifying batches which were opened to any worker were systematically reviewed. In contrast, only up to 10\% of the trusted worker annotations were randomly selected for manual review. The quality of the annotations was assessed based on the following criteria:

\begin{itemize}[noitemsep,topsep=0pt]
    \item \textit{completeness}: most of the differences between pose $A$ and pose $B$ were addressed in the annotation;
    \item \textit{direction accuracy}: the annotation explains how to go from pose $A$ to pose $B$, and not the reverse;
    \item \textit{left/right accuracy}: the words `\textit{left}' and `\textit{right}' were used in the body's frame of reference;
    \item \textit{3D consideration}: the annotation fits the 3D information, no guess was taken on occluded body parts, or ambiguous postures;
    \item \textit{no distance metric}: the annotation does not contain any distance metric (\eg, `\textit{one meter apart}'), which would not scale to bodies of different size;
    \item \textit{writing quality}: correct grammar and formulation.
\end{itemize}

\myparagraph{Length of the human-written annotations.} Figure~\ref{fig:length_hist} shows the length distribution of the collected annotations. We here refer to the length as the number of words, excluding punctuation. While the annotations were constrained to be at least 10-word long, they tend to count about 30 words, suggesting that the differences between two similar poses $A$ and $B$ are both subtle and several.

\subsection{Automatic annotations}

We explain here in more details the learning-free process to automatically generate modifiers. The different steps of the pipeline are illustrated in Figure~\ref{fig:auto_comparative_pipeline}. We comment on some of those steps.

\myparagraph{Code extraction.}
Two of the elementary paircodes are basically variation-versions of the initial posecodes~\cite{posescript}: we look at the change in angle posecode or distance posecode between pose $A$ and pose $B$. The third kind of paircode studies the variation in position of a keypoint along the x-, y- or z- axis. All three paircodes are computed on the orientation-normalized bodies, so that the produced instructions would not depend on the change in global orientation of the body between pose $A$ and pose $B$. This last part is treated separately, and yields a sentence that is added at the beginning of the modifier. \\
We also resort to the posecodes of both poses $A$ and $B$ to define super-paircodes, and thus gain in abstraction or formulation quality. There can be several ways to achieve the same paircode, each way comprising at least two conditions (posecode and paircode mixed together). Some posecodes of pose $B$, if statistically rare, are also included in the final modifier, \eg `\textit{the hands should be shoulder-width apart}', `\textit{the left thigh should be parallel with the ground}'. Posecodes of pose $A$ are only useful for super-paircode computations.

\myparagraph{Code selection and aggregation.}
We proceed as in ~\cite{posescript}. Trivial codes are removed. The codes (paircodes + posecodes) are aggregated based on simple syntactic rules depending on shared information between codes.

\myparagraph{Code ordering.}
The final set of codes is semantically ordered to produce modifiers that are easier to read and closer to what a human would write (\ie, describe about everything related to the right arm at once, instead of scattering pieces of information everywhere in the text). This step did not exist in the PoseScript automatic pipeline. Specifically, we design a directed graph where the nodes represent the body parts and the edges define a relation of inclusion or proximity between them (\eg \textit{torso}$\rightarrow$\textit{left shoulder}, \textit{arm}$\rightarrow$\textit{forearm}). For each pose pair, we perform a randomized depth walk through the graph: starting from the \textit{body} node, we choose one node at random among the ones directly accessible, then reiterate the process from that node until we reach a leaf; at that point, we come back to the last visited node leading to non-visited nodes and sample one child node at random. We use the order in which the body parts are visited to order the paircodes.

\myparagraph{Code conversion.}
Codes are converted to pieces of text by plugging information into a randomly chosen template sentence associated to each of them. The pieces of text are next concatenated thanks to transition texts. Verbs are conjugated accordingly to the chosen transition (\eg ``while + gerund'') and code (\eg posecodes lead to ``[...] should be'' sentences).

We refer to the code for the detailed and complete list of paircodes and super-paircodes definition.

\section{Original triplets of the generation examples}
\label{app:ground_truth}

In this section, we provide the original triplets for the generation results presented in Figure~\ref{fig:b_generation_results} (see Figure \ref{fig:b_generation_results_GT}) and in Figure~\ref{fig:text_generation_qualitatives} (see Figure \ref{fig:text_generation_qualitatives_GT}). While this ground truth may ease the comparison, it is not the only true answer for a generative model: multiple valid results could be produced. The GT was purposely omitted to prevent judgment bias, but is added here for reference.

\begin{figure*}[t]
\centering
\includegraphics[width=\textwidth]{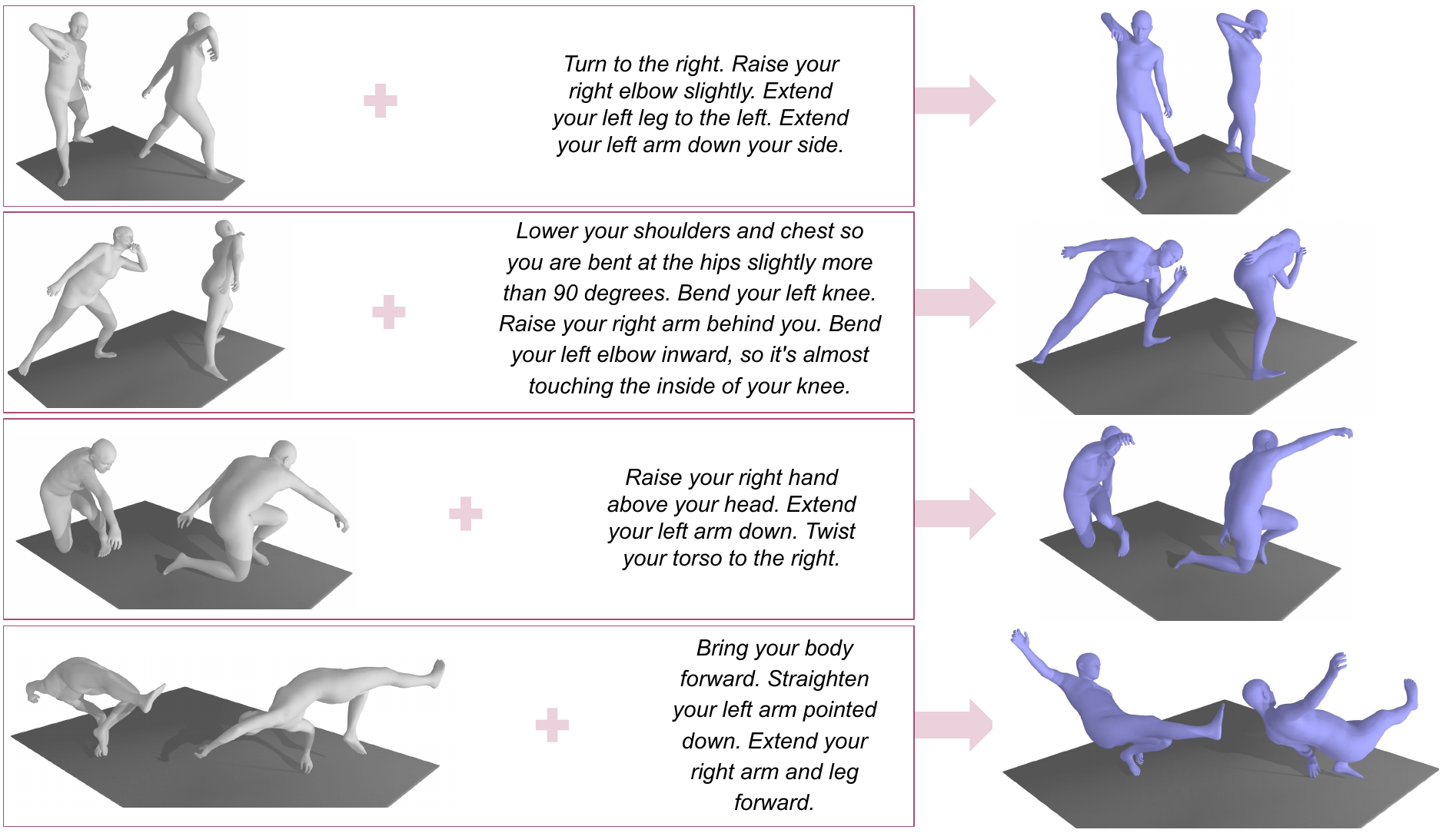} \\[-0.18cm]
\caption{\textbf{Original poses $B$} for the text-based pose editing task and \dname queries presented in Figure~\ref{fig:b_generation_results}. Two views of the each pose are shown on the same ground plane. Pose $A$ is shown in grey, pose $B$ in purple.}
\label{fig:b_generation_results_GT}
\vspace{-0.3cm}
\end{figure*}

\begin{figure*}[t]
\centering
   \includegraphics[width=\textwidth]{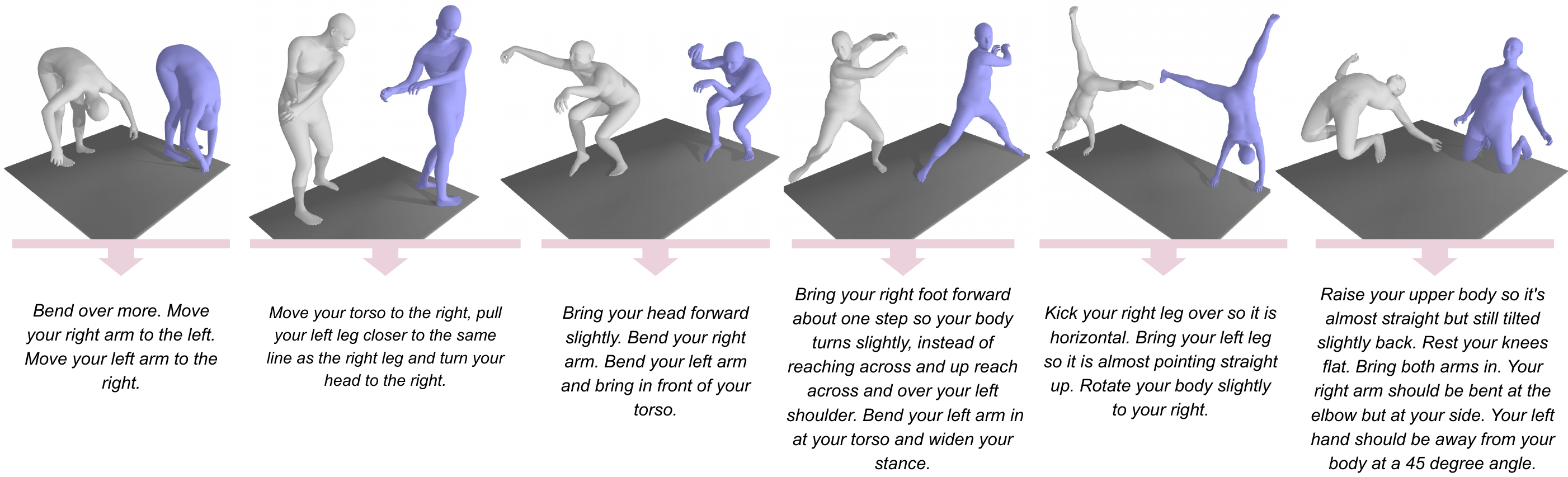} \\[-0.12cm]
\caption{\textbf{Original correctional feedback annotation} for \dname pose pairs presented in Figure~\ref{fig:text_generation_qualitatives_GT}. Pose $A$ is shown in grey, pose $B$ in purple.}
\label{fig:text_generation_qualitatives_GT}
\vspace{-0.3cm}
\end{figure*}
\section{Miscellaneous visualizations}
\label{app:misc_visu}

\myparagraph{Robot teaching application.}
The choice of modifiers in Natural Language to learn the difference between two poses proves especially useful in applications where direct manipulation is not possible, for instance in the case of robot teaching. Figure~\ref{fig:robot_application} shows a snapshot of a demo where a two-arm robot pose is optimized to match SMPL keypoints obtained from textual instructions.

\begin{figure}[t]
\centering
\includegraphics[width=\linewidth]{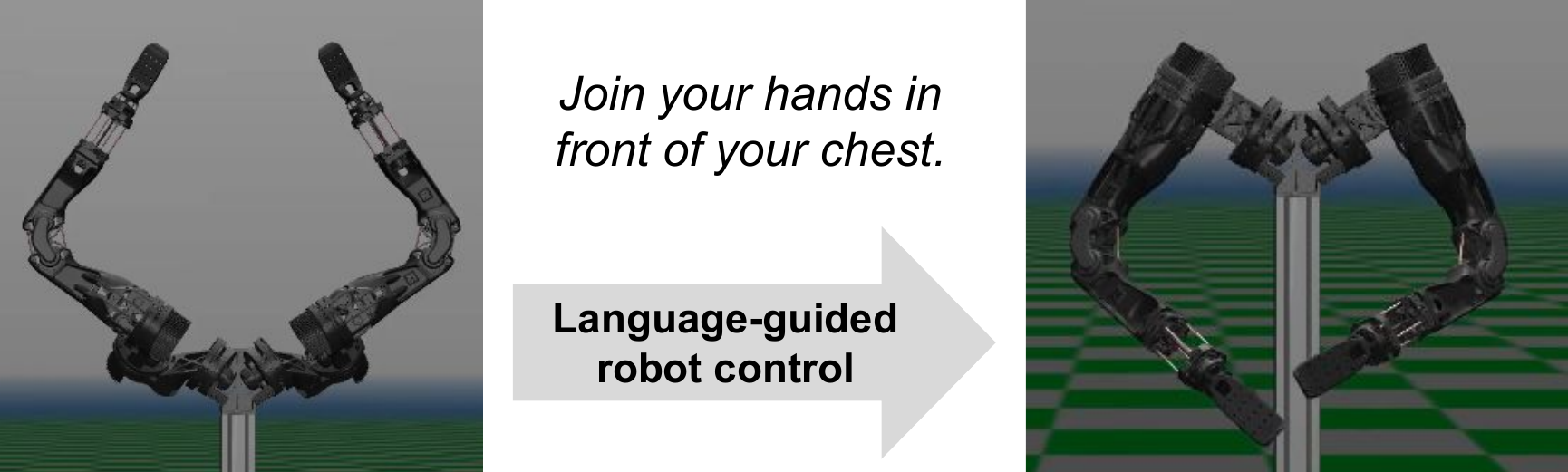}
\caption{\textbf{Robot teaching application.}}
\label{fig:robot_application}
\end{figure}

\myparagraph{The PoseCopy behavior.}
The PoseCopy setting for the text-based pose editing task consists in training the model with a proportion of the data where the text is emptied and pose $B$ becomes a copy-paste of pose $A$. This training configuration makes it possible for the model to yield the exact same pose as the initial one, when no correctional instruction is specified, see Figure~\ref{fig:posecopy} for an example. Besides, we hypothesize that this setting encourages the model to better pay attention to pose $A$.

\begin{figure}[t]
\centering
\includegraphics[width=\linewidth]{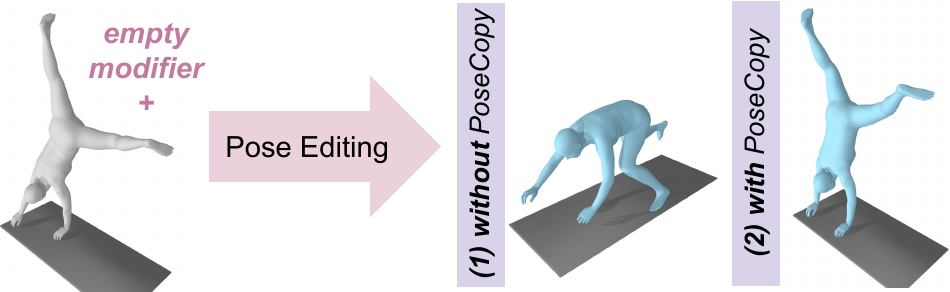}
\caption{\textbf{Effect of training with PoseCopy.}}
\label{fig:posecopy}
\end{figure}
\section{Implementation details}
\label{app:details}

\myparagraph{Architecture details.} We follow the VPoser~\cite{smplx} architecture for our pose encoder, modified to account for the 52 joints of the SMPL-H~\cite{mano} body model. In the `glove+bigru' configuration of our pose editing baseline, GloVe word embeddings are of size 300 and we use a bidirectional GRU with one layer and hidden state features of size 512. In the transformer configuration, we use a frozen pretrained DistilBERT model to encode the text tokens. The transformer afterwards is composed of 4 layers with 4 heads and feed-forward networks with 1024 dimensions. It relies on GELU~\cite{hendrycks2016gaussian-gelu} activations and uses a $0.1$ dropout. The text embedding is eventually obtained by performing an average pooling. The transformer in our correctional text generation baseline is the same as for pose editing, except that we use 8 heads. In our models for both tasks, the poses and texts are encoded in latent spaces of dimensions $d{=}32$ and $n{=}128$ ($n{=}512$ for the text generation task) respectively.

\myparagraph{Optimization and training details.} We trained our models with the Adam~\cite{kingma2014adam} optimizer, a batch size of 128, a learning rate of $10^{-5}$ ($10^{-4}$ for pretraining; and $10^{-6}$ for finetuning in the case of pose editing) and a weight decay of $10^{-4}$ ($10^{-5}$ for finetuning in the case of pose editing). The pose editing model was trained for 10,000 epochs (half for pretraining and half for finetuning, or 10,000 straight if no pretraining was involved), while the text generation model was trained for 3,000 epochs for pretraining and 2,000 for finetuning.
In the \textit{PoseCopy} setting, 50\% of the batch is randomly used in ``copy'' mode (\ie, empty text, with poses $A$ and $B$ being the same).

\myparagraph{Why using the ELBO metric?} The ELBO is well suited to VAEs~\cite{vae}: it balances reconstruction and KL into a lower bound on the data log likelihood, a universal quantity for comparing likelihood-based generative models. It accounts for the probabilistic nature of the model, by evaluating the target under the output distribution. In a VAE framework, reporting reconstruction errors only does not penalize the model for storing a lot of information in the latent variable produced by the encoder. The extreme case of an encoder that learns an identity function would appear optimal, yet fail at test time when the ground truth is no longer available for encoding. By contrast, the ELBO takes both reconstruction and the amount of information given by the encoder (the KL term) into account, and combines them into a lower bound on the data log likelihood.

\myparagraph{Hand data.} We used the hand data (fingers joints) for all ours experiments, but note that this was not necessary, given that the hands all have the same pose for \dname human-annotated pose pairs. In case more data with relevant hand information is annotated in the future, we suggest to keep the original hand data for the pairs annotated in this version of the dataset, as some annotators may have referred to them in their instructions.

\end{document}